\documentclass[10pt,twocolumn,letterpaper]{article}

\usepackage{cvpr}
\usepackage{times}
\usepackage{epsfig}
\usepackage{graphicx}
\usepackage{amsmath}
\usepackage{amssymb}

\usepackage{booktabs}

\usepackage{pgfplots} %
\usepackage{pgfplotstable} %
\pgfplotsset{compat=newest} %
\usetikzlibrary{plotmarks} %
\usetikzlibrary{colorbrewer} %

\usepackage[pagebackref=true,breaklinks=true,letterpaper=true,colorlinks,bookmarks=false]{hyperref}

\def\eg{\emph{e.g}\onedot} 
\def\ie{\emph{i.e}\onedot} 
 
 \def\vs{\emph{vs}\onedot}
\def\wrt{w.r.t\onedot} 
\def\etal{\emph{et al}\onedot}

\newcommand{\figref}[1]{Fig\onedot~\ref{#1}}

\newcommand{\tabref}[1]{Tab\onedot~\ref{#1}}

\usepackage{pifont}
\newcommand{\cmark}{\ding{51}}

\cvprfinalcopy %

\def\cvprPaperID{784} %

\begin{document}

\title{Panoptic-DeepLab:\\A Simple, Strong, and Fast Baseline for Bottom-Up Panoptic Segmentation}

\author{
Bowen Cheng$^{1,2}$, Maxwell D.~Collins$^{2}$, Yukun Zhu$^{2}$, Ting Liu$^{2}$,\\
Thomas S. Huang$^{1}$, Hartwig Adam$^{2}$, Liang-Chieh Chen$^{2}$\\
\\
{$^1$UIUC \hspace{2mm} $^2$Google Research}
}

\maketitle

\begin{abstract}
In this work, we introduce Panoptic-DeepLab, a simple, strong, and fast system for panoptic segmentation, aiming to establish a solid baseline for bottom-up methods that can achieve comparable performance of two-stage methods while yielding fast inference speed. In particular, Panoptic-DeepLab adopts the dual-ASPP and dual-decoder structures specific to semantic, and instance segmentation, respectively. The semantic segmentation branch is the same as the typical design of any semantic segmentation model (\eg, DeepLab), while the instance segmentation branch is class-agnostic, involving a simple instance center regression. As a result, our single Panoptic-DeepLab simultaneously ranks first at all three Cityscapes benchmarks, setting the new state-of-art of 84.2\% mIoU, 39.0\% AP, and 65.5\% PQ on test set. Additionally, equipped with MobileNetV3, Panoptic-DeepLab runs nearly in real-time with a single $1025\times2049$ image (15.8 frames per second), while achieving a competitive performance on Cityscapes (54.1 PQ\% on test set). On Mapillary Vistas test set, our ensemble of six models attains 42.7\% PQ, outperforming the challenge winner in 2018 by a healthy margin of 1.5\%. Finally, our Panoptic-DeepLab also performs on par with several top-down approaches on the challenging COCO dataset. For the first time, we demonstrate a bottom-up approach could deliver state-of-the-art results on panoptic segmentation. 
\end{abstract}

\section{Introduction}
\label{sec:intro}

\begin{figure}[!t]
    \centering
    \includegraphics[width=0.46\textwidth]{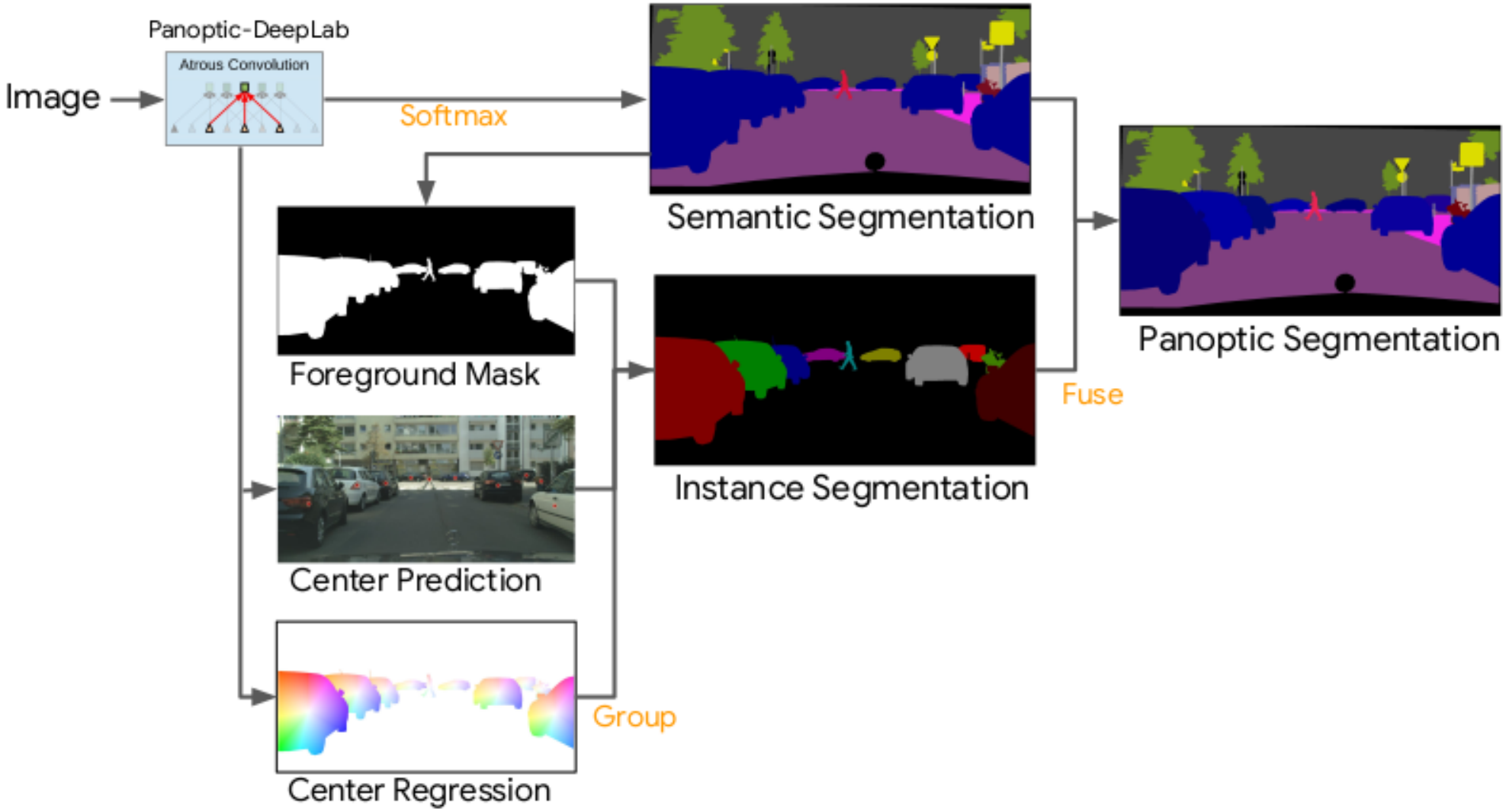}
    \caption{Our Panoptic-DeepLab predicts three outputs: semantic segmentation, instance center prediction and instance center regression. Class-agnostic instance segmentation, obtained by grouping predicted foreground pixels to their closest predicted instance centers, is then fused with semantic segmentation by majority-vote rule to generate final panoptic segmentation.}
    \label{fig:teaser}
\end{figure}

Panoptic segmentation, unifying semantic segmentation and instance segmentation, has received a lot of attention thanks to the recently proposed panoptic quality metric \cite{kirillov2018panoptic} and associated recognition challenges \cite{lin2014microsoft,cordts2016cityscapes,neuhold2017mapillary}. The goal of panoptic segmentation is to assign a unique value, encoding both semantic label and instance id, to every pixel in an image.
It requires identifying the class and extent of each individual `thing' in the image, and labelling all pixels that belong to each `stuff' class.

The task of panoptic segmentation introduces challenges that preceding methods are unsuited to solve. Models typically used in the separate instance and semantic segmentation literature have diverged, and fundamentally different approaches dominate in each setting.
For panoptic segmentation, the top-down methods \cite{xiong2019upsnet, kirillov2019panoptic, li2018learning, li2018attention, porzi2019seamless}, attaching another semantic segmentation branch to Mask R-CNN \cite{he2017mask}, generate overlapping instance masks as well as duplicate pixel-wise semantic predictions. To settle the conflict, the commonly employed heuristic resolves overlapping instance masks by their predicted confidence scores \cite{kirillov2018panoptic}, or even by the pairwise relationship between categories \cite{li2018attention} (\eg, {\it ties} should be always in front of {\it person}). Additionally, the discrepancy between semantic and instance segmentation results are sorted out by favoring the instance predictions. Though effective, it may be hard to implement the hand-crafted heuristics in a fast and parallel fashion. Another effective way is to develop advanced modules to fuse semantic and instance segmentation results \cite{li2018attention,li2018learning,xiong2019upsnet}. However, these top-down methods are usually slow in speed, resulted from the multiple sequential processes in the pipeline.

On the other hand, bottom-up methods naturally resolve the conflict by predicting non-overlapping segments. Only few works \cite{yang2019deeperlab,gao2019ssap} adopt the bottom-up approach, which typically starts with a semantic segmentation prediction followed by grouping operations to generate instance masks. Tackling panoptic segmentation in such a sequential order allows a simple and fast scheme, such as majority vote \cite{yang2019deeperlab}, to merge semantic and instance segmentation results. Although obtaining promising fast inference speed, bottom-up approaches still demonstrate inferior performance compared to top-down ones prevailing in public benchmarks \cite{lin2014microsoft,cordts2016cityscapes,neuhold2017mapillary}.

The difficulties faced by top-down methods, and the dearth of previous
investigations into complementary approaches motivate us to establish a simple, strong, and fast bottom-up baseline for panoptic segmentation. Our proposed {\bf Panoptic-DeepLab} (\figref{fig:teaser}) requires only three loss functions during training, and introduces extra marginal parameters as well as additional slight computation overhead when building on top of a modern semantic segmentation model. The design of the proposed Panoptic-DeepLab is conceptually simple, adopting dual-ASPP and dual-decoder modules specific to semantic segmentation and instance segmentation, respectively. The semantic segmentation branch follows the typical design of any semantic segmentation model (\eg, DeepLab \cite{deeplabv3plus2018}), while the instance segmentation branch involves a simple instance center regression \cite{ballard1981generalizing,kendall2018multi}, where the model learns to predict instance centers as well as the offset from each pixel to its corresponding center, resulting in an extremely simple grouping operation by assigning pixels to their closest predicted center. Additionally, with fast GPU implementation of the merging operation, Panoptic-DeepLab delivers near real-time end-to-end panoptic segmentation prediction. 

We conduct experiments on several popular panoptic segmentation datasets. On Cityscapes test set \cite{cordts2016cityscapes}, a {\it single} Panoptic-DeepLab model (without fine-tuning on different tasks) achieves state-of-the-art performance of 65.5\% PQ, 39.0\% AP, and 84.2\% mIoU, simultaneously ranking first on {\it all} three Cityscapes tasks when comparing with published works. On Mapillary Vistas \cite{neuhold2017mapillary}, our best {\it single} model attains 40.6\% PQ on val set, while employing an ensemble of 6 models reaches a performance of 42.2\% PQ on val set and 42.7\% PQ on test set, outperforming the winner of Mapillary Vistas Panoptic Segmentation Challenge in 2018 by a healthy margin of 1.5\% PQ. 
For the first time, we show a bottom-up approach could deliver state-of-the-art panoptic segmentation results on both Cityscapes and Mapillary Vistas. On COCO \cite{lin2014microsoft} test-dev set, our Panoptic-DeepLab also demonstrates state-of-the-art results, performing on par with several top-down approaches. Finally, we provide extensive experimental results and disclose every detail in our system. We hope our Panoptic-DeepLab could serve as a solid baseline to facilitate the research on panoptic segmentation, especially from the bottom-up perspective.
\section{Related Works}
\begin{figure*}[!t]
    \centering
    \includegraphics[width=1.0\textwidth]{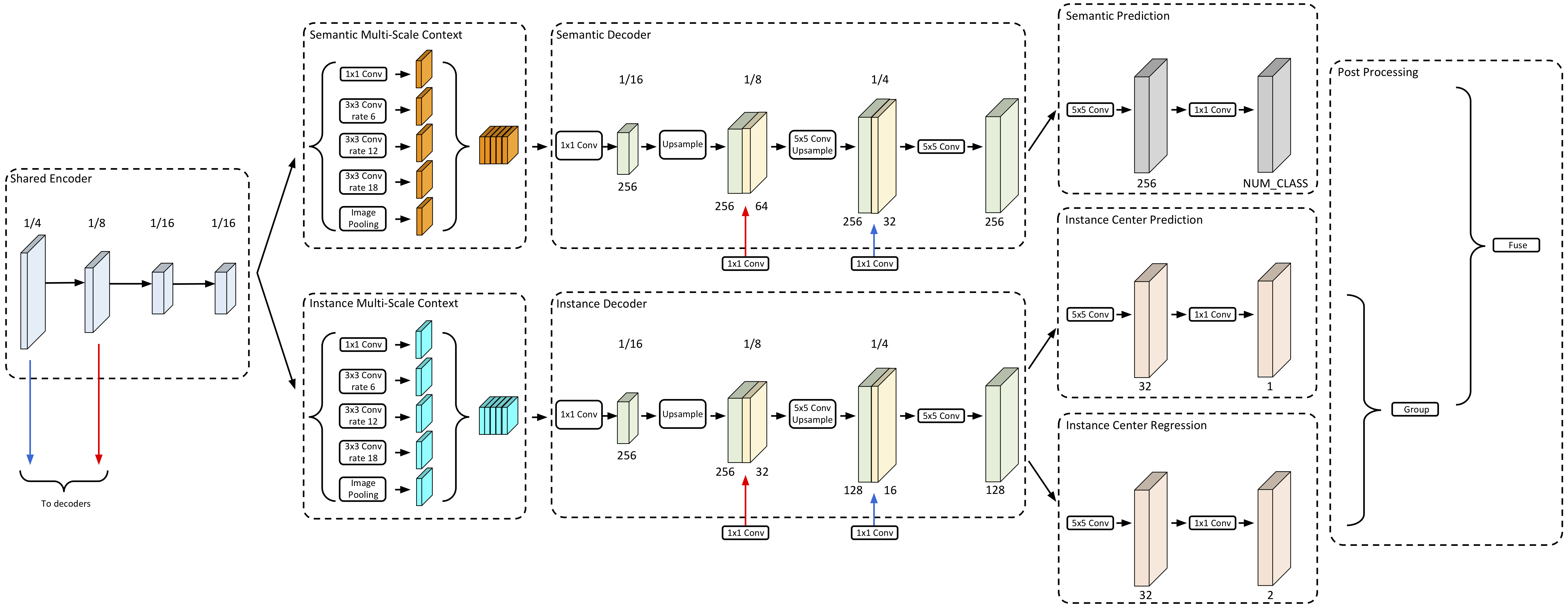}
    \caption{Our Panoptic-DeepLab adopts dual-context and dual-decoder modules for semantic segmentation and instance segmentation predictions. We apply atrous convolution in the last block of a network backbone to extract denser feature map. The Atrous Spatial Pyramid Pooling (ASPP) is employed in the context module as well as a light-weight decoder module consisting of a single convolution during each upsampling stage. The instance segmentation prediction is obtained by predicting the object centers and regressing every foreground pixel (\ie, pixels with predicted `thing` class) to their corresponding center. The predicted semantic segmentation and class-agnostic instance segmentation are then fused to generate the final panoptic segmentation result by the "majority vote" proposed by DeeperLab.}
    \label{fig:network_architecture}
    \vspace{-2mm}
\end{figure*}
We categorize current panoptic segmentation methods \cite{kirillov2018panoptic} into two groups: top-down and bottom-up approaches.

{\bf Top-down:} Most state-of-the-art methods tackle panoptic segmentation from the top-down or proposal-based perspective. These methods are often referred to as two-stage methods because they require an additional stage to generate proposals. Specifically, Mask R-CNN \cite{he2017mask} is commonly deployed to extract {\it overlapping instances}, followed by some post-processing methods to resolve mask overlaps. The remaining regions are then filled by a light-weight stuff segmentation branch. For example, TASCNet \cite{li2018learning} learns a binary mask to enforce the consistency between `thing' and `stuff' predictions. Liu \etal \cite{liu2019e2e} propose the Spatial Ranking module to resolve the overlapping instance masks. AUNet \cite{li2018attention} introduces attention modules to guide the fusion between `thing' and `stuff' segmentation. Panoptic FPN \cite{kirillov2019panoptic} endows Mask R-CNN \cite{he2017mask} with a semantic segmentation branch. UPSNet \cite{xiong2019upsnet} develops a parameter-free panoptic head which resolves the conflicts in `thing'-`stuff' fusion by predicting an extra unknown class. Porzi \etal \cite{porzi2019seamless} integrate the multi-scale features from FPN \cite{lin2017feature} with a light-weight DeepLab-inspired module \cite{chen2018deeplabv2}. AdaptIS \cite{sofiiuk2019adaptis} generates instance masks with point proposals.

{\bf Bottom-up:} On the other hand, there are few bottom-up or proposal-free methods for panoptic segmentation. These works typically get the semantic segmentation prediction before detecting instances by grouping `thing' pixels into clusters. The first bottom-up approach, DeeperLab \cite{yang2019deeperlab}, adopts bounding box corners as well as object centers for class-agnostic instance segmentation, coupled with DeepLab semantic segmentation outputs \cite{deeplabv12015,chen2017deeplabv3}. Recently, SSAP \cite{gao2019ssap} proposes to group pixels based on a pixel-pair affinity pyramid \cite{liu2018affinity} with an efficient graph partition method \cite{keuper2015efficient}. Unfortunately, given its simplicity (\ie, a single pass of the system for prediction), bottom-up approaches perform inferiorly to top-down methods at almost {\it all} public benchmarks. In this work, we aim to push the envelope of bottom-up approaches. We note that there are several instance segmentation works \cite{zhang2015monocular,uhrig2016pixel,zhang2016instance,bai2017deep,liu2017sgn,kirillov2017instancecut,newell2017associative,fathi2017semantic,de2017semantic,liang2018proposal,kulikov2018instance,kendall2018multi,liu2018affinity,neven2019instance,yolact}, which could be potentially extended to bottom-up panoptic segmentation. Additionally, our method bears a similarity to Hough-Voting-based methods \cite{ballard1981generalizing,leibe2004combined,gall2009class,barinova2012detection} and recent works by Kendall \etal \cite{kendall2018multi}, Uhrig \etal \cite{uhrig2018box2pix} and Neven \etal \cite{neven2019instance} in the sense that our class-agnostic instance segmentation is obtained by regressing foreground pixels to their centers. However, our method is even simpler than theirs: we directly predict the instance center locations and group pixels to their closest predicted centers. As a result, our method does not require the clustering method OPTICS \cite{ankerst1999optics} used in \cite{kendall2018multi}, or the advanced clustering loss function proposed in \cite{neven2019instance}. Finally, our model employs the parallel multi-head prediction framework similar to \cite{uhrig2016pixel,kokkinos2017ubernet,neven2017fast}.

{\bf Keypoint representation:} Recently, keypoint representations have been used for instance segmentation and object detection. Newell \etal \cite{newell2017associative} group pixels by embedding vectors. PersonLab \cite{papandreou2018personlab} generates person segmentation masks and groups them into instances by learning offset to their detected keypoints. CornerNet \cite{Law_2018_ECCV} detects objects by predicting paired corners and group corners based on \cite{newell2017associative}. ExtremeNet \cite{zhou2019bottom} groups `extreme points' \cite{papadopoulos2017extreme} according to the relation to a center point. Zhou \etal \cite{Zhou2019Objects} and Duan \etal \cite{Duan2019CenterNet} exploit instance centers for object detection. Following the same direction, we represent each instance by its center and take a step further by showing that such a simple representation is able to achieve state-of-the-art panoptic segmentation results on several challenging datasets. Different from keypoint-based detection, our Panoptic-DeepLab only requires class-agnostic object center prediction.

\section{Panoptic-DeepLab}
\label{sec:methods}

As shown in \figref{fig:network_architecture}, our proposed Panoptic-DeepLab is deployed in a bottom-up and single-shot manner. 

\subsection{Architecture}
Panoptic-DeepLab consists of four components: (1) an encoder backbone shared for both semantic segmentation and instance segmentation, (2) decoupled ASPP modules and (3) decoupled decoder modules specific to each task, and (4) task-specific prediction heads.

{\bf Basic architecture:} The encoder backbone is adapted from an ImageNet-pretrained neural network paired with atrous convolution for extracting denser feature maps in its last block. Motivated by~\cite{cheng2018revisiting,cheng2018decoupled,neven2019instance}, we employ separate ASPP and decoder modules for semantic segmentation and instance segmentation, respectively, based on the hypothesis that those two branches requires different contextual and decoding information, which is empirically verified in the following section. Our light-weight decoder module follows DeepLabV3+~\cite{deeplabv3plus2018} with two modifications: (1) we introduce an additional low-level feature with output stride 8 to the decoder, thus the spatial resolution is gradually recovered by a factor of 2, and (2) in each upsampling stage we apply a {\it single} $5\times5$ depthwise-separable convolution \cite{howard2017mobilenets}.

{\bf Semantic segmentation head:} We employ the weighted bootstrapped cross entropy loss, proposed in \cite{yang2019deeperlab}, for semantic segmentation, predicting both `thing' and `stuff' classes. The loss improves over bootstrapped cross entropy loss \cite{wu2016bridging, bulo2017loss, pohlen2016full} by weighting each pixel differently.

{\bf Class-agnostic instance segmentation head:} Motivated by Hough Voting  \cite{ballard1981generalizing,kendall2018multi}, we represent each object instance by its center of mass. For every foreground pixel (\ie, pixel whose class is a `thing'), we further predict the offset to its corresponding mass center. During training, groundtruth instance centers are encoded by a 2-D Gaussian with standard deviation of 8 pixels \cite{tompson2014joint}.  In particular, we adopt the Mean Squared Error (MSE) loss to minimize the distance between predicted heatmaps and 2D Gaussian-encoded groundtruth heatmaps. We use $L_1$ loss for the offset prediction, which is only activated at pixels belonging to object instances. During inference, predicted foreground pixels (obtained by filtering out background `stuff' regions from semantic segmentation prediction) are grouped to their closest predicted mass center, forming our class-agnostic instance segmentation results, as detailed below.

\subsection{Panoptic Segmentation}
During inference, we use an extremely simple grouping operation to obtain instance masks, and a highly efficient majority voting algorithm to merge semantic and instance segmentation into final panoptic segmentation.

{\bf Simple instance representation:} We simply represent each object by its center of mass, $\{\mathcal{C}_{n}: (i_{n}, j_{n})\}$. To obtain the center point prediction, we first perform a keypoint-based non-maximum suppression (NMS) on the instance center heatmap prediction, essentially equivalent to applying max pooling on the heatmap prediction and keeping locations whose values do not change before and after max pooling. Finally, a hard threshold is used to filter out predictions with low confidence, and only locations with top-k highest confidence scores are kept. In experiments, we use max-pooling with kernel size 7, threshold 0.1, and $k=200$.

{\bf Simple instance grouping:} To obtain the instance id for each pixel, we use a simple instance center regression. For example, consider a predicted `thing' pixel at location $(i, j)$, we predict an offset vector $\mathcal{O}(i,j)$ to its instance center. $\mathcal{O}(i,j)$ is a vector with two elements, representing the offset in horizontal and vertical directions, respectively. The instance id for the pixel is thus the index of the closest instance center after moving the pixel location $(i,j)$ by the offset $\mathcal{O}(i,j)$. That is, $$\hat{k}_{i,j}=\operatorname*{argmin}_k ||\mathcal{C}_k-((i, j) + \mathcal{O}(i,j))||^2$$
where $\hat{k}_{i,j}$ is the predicted instance id for pixel at $(i, j)$.

We use semantic segmentation prediction to filter out `stuff' pixels whose instance id are always set to 0.

{\bf Efficient merging:} Given the predicted semantic segmentation and class-agnostic instance  segmentation results, we adopt a fast and parallelizable method to merge the results, following the ``majority vote'' principle proposed in DeeperLab \cite{yang2019deeperlab}. In particular, the semantic label of a predicted instance mask is inferred by the majority vote of the corresponding predicted semantic labels. This operation is essentially accumulating the class label histograms, and thus is  efficiently implemented in GPU, which takes only 3 ms when operating on a $1025\times2049$ input.

\subsection{Instance Segmentation}
Panoptic-DeepLab can also generate instance segmentation predictions as a by-product. To properly evaluate the instance segmentation results, one needs to associate a confidence score with each predicted instance mask. Previous bottom-up instance segmentation methods use some heuristics to obtain the confidence scores. For example, DWT \cite{bai2017deep} and SSAP \cite{gao2019ssap} use an average of semantic segmentation scores for some easy classes and use random scores for other harder classes. Additionally, they remove masks whose areas are below a certain threshold for each class. On the other hand, our Panoptic-DeepLab does not adopt any heuristic or post processing for instance segmentation. Motivated by YOLO \cite{redmon2016you}, we compute the class-specific confidence score for each instance mask as 

$$Score(Objectness) \times Score(Class)$$
where $Score(Objectness)$ is unnormalized objectness score obtained from the class-agnostic center point heatmap, and $Score(Class)$ is obtained from the average of semantic segmentation predictions within the predicted mask region.

\section{Experiments}

\begin{table*}[!t]
  \centering
  \scalebox{0.75}{
  \begin{tabular}{ c  c  c  c  c  c  c  c | c | c | c | c | c}
    \toprule[0.2em]
    Adam & MSE & De. x2 & ASPP x2  & L-Crop & $\text{C}_{\text{Sem}}=256$ & $\text{C}_{\text{Ins}}=256$ & Sem. Only & PQ (\%) & AP (\%) & mIoU (\%) & Params (M) & M-Adds (B) \\
    \toprule[0.2em]
    & & & & & & & & 60.3 & 32.7 & 78.2 & 41.85 & 496.84\\
    \cmark & & & & & & & & 61.0 & 34.3 & 79.4 & 41.85 & 496.84\\
    \cmark & \cmark & & & & & & & 61.8 & 33.8 & 78.6 & 41.85 & 496.84\\
    \cmark & \cmark & \cmark & & & & & & 60.8 & 32.7 & 79.0 & 41.93 & 501.88\\
    \cmark & \cmark & \cmark & \cmark & & & & & 62.5 & 33.9 & 78.7 & 43.37 & 517.17\\
    \cmark & \cmark & \cmark & \cmark & \cmark & & & & 62.7 & 34.5 & 79.6 & 43.37 & 517.17\\
    \cmark & \cmark & \cmark & \cmark & \cmark & \cmark & & & \textbf{63.0} & \textbf{35.3} & \textbf{80.5} & 46.72 & 547.49\\
    \cmark & \cmark & \cmark & \cmark & \cmark & \cmark & \cmark & & 62.1 & 35.1 & 80.3 & 46.88 & 573.86\\
    \midrule\midrule
    \cmark & & & & \cmark & \cmark & & \cmark & - & - & 80.3 & 43.60 & 518.84\\
    \bottomrule[0.1em]
  \end{tabular}
  }
  \caption{Ablation studies on Cityscapes {\it val} set. {\bf Adam}: Adam optimizer. {\bf MSE}: MSE loss for instance center. {\bf De. x2}: Dual decoder. {\bf ASPP x2}: Dual ASPP. {\bf L-Crop}: Large crop size. {\bf $\text{C}_{\text{Sem}}=256$}: 256 (instead of 128) channels in semantic segmentation branch. {\bf $\text{C}_{\text{Ins}}=256$}: 256 (instead of 128) channels in instance segmentation branch. {\bf Sem. Only}: Only semantic segmentation. M-Adds are measured \wrt a $1025\times2049$ input.}
  \label{tab:cityscapes_ablation}
  \vspace{-2mm}
\end{table*}

\begin{table}[!t]
  \centering
  \scalebox{0.7}{
  \begin{tabular}{c | c | c | c | c | c | c  }
    \toprule[0.2em]
    Method & Extra Data & Flip & MS  & PQ (\%) & AP (\%) & mIoU (\%) \\
    \toprule[0.2em]
    \multicolumn{7}{c}{w/o Extra Data}\\
    \midrule
    TASCNet~\cite{li2018learning} & & & & 55.9 & - & - \\
    Panoptic FPN~\cite{kirillov2019panoptic} & & & & 58.1 & 33.0 & 75.7 \\
    AUNet~\cite{li2018attention} & & & & 59.0 & 34.4 & 75.6 \\
    UPSNet~\cite{xiong2019upsnet} & & & & 59.3 & 33.3 & 75.2 \\
    UPSNet~\cite{xiong2019upsnet} & & \cmark & \cmark & 60.1 & 33.3 & 76.8 \\
    Seamless~\cite{porzi2019seamless} & & & & 60.3 & 33.6 & 77.5 \\
    AdaptIS~\cite{sofiiuk2019adaptis} & & \cmark & & 62.0 & 36.3 & 79.2 \\
    \midrule
    DeeperLab~\cite{yang2019deeperlab} & & & & 56.5 & - & - \\
    SSAP~\cite{gao2019ssap} & & \cmark & \cmark & 61.1 & 37.3 & - \\
    \midrule\midrule
    Panoptic-DeepLab & & & & 63.0 & 35.3 & 80.5 \\
    Panoptic-DeepLab & & \cmark & & 63.4 & 36.1 & 80.9 \\
    Panoptic-DeepLab & & \cmark & \cmark & \textbf{64.1} & \textbf{38.5} & \textbf{81.5} \\
    \midrule
    \multicolumn{7}{c}{w/ Extra Data}\\
    \midrule
    TASCNet~\cite{li2018learning} & COCO & & & 59.3 & 37.6 & 78.1 \\
    TASCNet~\cite{li2018learning} & COCO & \cmark & \cmark & 60.4 & 39.1 & 78.7 \\
    UPSNet~\cite{xiong2019upsnet} & COCO & & & 60.5 & 37.8 & 77.8 \\
    UPSNet~\cite{xiong2019upsnet} & COCO & \cmark & \cmark & 61.8 & 39.0 & 79.2 \\
    Seamless~\cite{porzi2019seamless} & MV &  &  & 65.0 & - & 80.7 \\
    \midrule\midrule
    Panoptic-DeepLab & MV & & & 65.3 & 38.8 & 82.5 \\
    Panoptic-DeepLab & MV & \cmark & & 65.6 & 39.4 & 82.6 \\
    Panoptic-DeepLab & MV & \cmark & \cmark & \textbf{67.0} & \textbf{42.5} & \textbf{83.1} \\
    \bottomrule[0.1em]
  \end{tabular}
  }
  \caption{Cityscapes {\it val} set. {\bf Flip:} Adding left-right flipped inputs. {\bf MS:} Multiscale inputs. {\bf MV:} Mapillary Vistas.}
  \label{tab:cityscapes_val}
  \vspace{-2mm}
\end{table}

\begin{table}[!t]
  \centering
  \scalebox{0.77}{
  \begin{tabular}{c | c | c | c | c  }
    \toprule[0.2em]
    Method & Extra Data & PQ (\%) & AP (\%) & mIoU (\%) \\
    \toprule[0.2em]
    \multicolumn{5}{c}{Semantic Segmentation}\\
    \midrule
    GFF-Net~\cite{li2019gff} &  & - & - & 82.3 \\
    Zhu~\etal~\cite{zhu2019improving} & C, V, MV & - & - & 83.5 \\
    Hyundai Mobis AD Lab & C, MV & - & - & 83.8 \\
    \midrule
    \multicolumn{5}{c}{Instance Segmentation}\\
    \midrule
    AdaptIS~\cite{sofiiuk2019adaptis} & & - & 32.5 & - \\
    UPSNet~\cite{xiong2019upsnet} & COCO & - & 33.0 & - \\
    PANet~\cite{liu2018path} & COCO & - & 36.4 & - \\
    Sogou\_MM & COCO & - & 37.2 & - \\
    iFLYTEK-CV & COCO & - & 38.0 & - \\
    NJUST & COCO & - & 38.9 & - \\
    AInnoSegmentation & COCO & - & \textbf{39.5} & - \\
    \midrule
    \multicolumn{5}{c}{Panoptic Segmentation}\\
    \midrule
    SSAP~\cite{gao2019ssap} & & 58.9 & 32.7 & - \\
    TASCNet~\cite{li2018learning} & COCO & 60.7 & - & - \\
    Seamless~\cite{porzi2019seamless} & MV & 62.6 & - & - \\
    \midrule\midrule
    Panoptic-DeepLab & & 62.3 & 34.6 & 79.4 \\
    Panoptic-DeepLab & MV & \textbf{65.5} & 39.0 & \textbf{84.2} \\
    \bottomrule[0.1em]
  \end{tabular}
  }
  \caption{Cityscapes {\it test} set. {\bf C:} Cityscapes coarse annotation. {\bf V:} Cityscapes video. {\bf MV:} Mapillary Vistas.}
  \label{tab:cityscapes_test}
  \vspace{-2mm}
\end{table}

{\bf Cityscapes \cite{cordts2016cityscapes}:} The dataset consists of 2975, 500, and 1525 traffic-related images for training, validation, and testing, respectively. It contains 8 `thing' and 11 `stuff' classes.

{\bf Mapillary Vistas \cite{neuhold2017mapillary}:} A large-scale traffic-related dataset, containing 18K, 2K, and 5K images for training, validation and testing, respectively. It contains 37 `thing' classes and 28 `stuff' classes in a variety of image resolutions, ranging from $1024\times768$ to more than $4000\times6000$

{\bf COCO \cite{lin2014microsoft}:} There are 118K, 5K, and 20K images for training, validation, and testing, respectively. The dataset consists of 80 `thing` and 53 `stuff` classes.

{\bf Experimental setup:} We report mean IoU, average precision (AP), and panoptic quality (PQ) to evaluate the semantic, instance, and panoptic segmentation results.

All our models are trained using TensorFlow on 32 TPUs. We adopt a similar training protocol as in~\cite{deeplabv3plus2018}. In particular, we use the `poly' learning rate policy~\cite{liu2015parsenet} with an initial learning rate of $0.001$, fine-tune the batch normalization~\cite{ioffe2015batch}
parameters, perform random scale data augmentation during training, and optimize with Adam \cite{kingma2015adam} {\it without weight decay}. On Cityscapes, our best setting is obtained by training with whole image (\ie, crop size equal to $1025\times2049$) with batch size 32. On Mapillary Vistas, we resize the images to 2177 pixels at the longest side to handle the large input variations, and randomly crop $1025\times1025$ patches during training with batch size 64. %
On COCO, we resize the images to 1025 pixels at the longest side and train our models with crop size $1025\times1025$ with batch size 64. We set training iterations to 60K, 150K, and 200K for Cityscapes,  Mapillary Vistas, and COCO, respectively. During evaluation, due to the sensitivity of PQ \cite{xiong2019upsnet,li2018learning,porzi2019seamless}, we re-assign to `VOID' label all `stuff' segments whose areas are smaller than a threshold. The thresholds on Cityscapes, Mapillary Vistas, and COCO are 2048, 4096, and 4096, respectively. Additionally, we adopt multi-scale inference (scales equal to $\{0.5, 0.75, 1, 1.25, 1.5, 1.75, 2\}$ for Cityscapes and Mapillary Vistas and $\{0.5, 0.75, 1, 1.25, 1.5\}$ for COCO) and left-right flipped inputs, to further improve the performance. For all the reported results, unless specified, Xception-71 \cite{chollet2016xception, dai2017coco,deeplabv3plus2018} is employed as the backbone.

Panoptic-DeepLab is trained with three loss functions: weighted bootstrapped cross entropy loss for semantic segmentation head ($\mathcal{L}_{sem}$) \cite{yang2019deeperlab}; MSE loss for center heatmap head ($\mathcal{L}_{heatmap}$) \cite{tompson2014joint}; and L1 loss for center offset head ($\mathcal{L}_{offset}$) \cite{papandreou2018personlab}. The final loss $\mathcal{L}$ is computed as follows.

$$\mathcal{L} = \lambda_{sem}\mathcal{L}_{sem} + \lambda_{heatmap}\mathcal{L}_{heatmap} + \lambda_{offset}\mathcal{L}_{offset}$$
Specifically, we set $\lambda_{sem}=3$ for pixels belonging to instances with an area smaller than $64\times64$ and $\lambda_{sem}=1$ everywhere else, following DeeperLab~\cite{yang2019deeperlab}. To make sure the losses are in the similar magnitude, we set $\lambda_{heatmap}=200$ and $\lambda_{offset}=0.01$.

\subsection{Ablation Studies}

We conduct ablation studies on Cityscapes validation set, as shown in \tabref{tab:cityscapes_ablation}. Replacing SGD momentum optimizer with Adam optimizer yields 0.7\% PQ improvement. Instead of using the sigmoid cross entropy loss for training the heatmap (\ie, instance center prediction), it brings 0.8\% PQ improvement by applying the Mean Squared Error (MSE) loss to minimize the distance between the predicted heatmap and the 2D Gaussian-encoded groundtruth heatmap. It is more effective to adopt both dual-decoder and dual-ASPP, which gives us 0.7\% PQ improvement while maintaining similar AP and mIoU. Employing a large crop size $1025\times2049$ (instead of $513\times1025$) during training further improves the AP and mIoU by 0.6\% and 0.9\% respectively. Finally, increasing the feature channels from 128 to 256 in the semantic segmentation branch achieves our best result of 63.0\% PQ, 35.3\% AP, and 80.5\% mIoU.

{\bf Multi-task learning:} For reference, we train a Semantic-DeepLab under the same setting as the best Panoptic-DeepLab (last row of \tabref{tab:cityscapes_ablation}), showing that multi-task learning does not bring extra gain to mIoU. Note that Panoptic-DeepLab adds marginal parameters and small computation overhead over Semantic-DeepLab.

\subsection{Cityscapes}
{\bf Val set:} In \tabref{tab:cityscapes_val}, we report our Cityscapes validation set results. When using only Cityscapes {\it fine} annotations, our best Panoptic-DeepLab, with multi-scale inputs and left-right flips, outperforms the best bottom-up approach, SSAP, by 3.0\% PQ and 1.2\% AP, and is better than the best proposal-based approach, AdaptIS, by 2.1\% PQ, 2.2\% AP, and 2.3\% mIoU. When using extra data, our best Panoptic-DeepLab outperforms UPSNet by 5.2\% PQ, 3.5\% AP, and 3.9\% mIoU, and Seamless by 2.0\% PQ and 2.4\% mIoU. Note that we do not exploit any other data, such as COCO, Cityscapes {\it coarse} annotations, depth, or video.

{\bf Test set:} On the test set, we additionally employ the trick proposed in \cite{deeplabv3plus2018} that applies atrous convolution in the last two blocks within the backbone, with rate 2 and 4 respectively, during inference. This trick brings an extra 0.4\% AP and 0.2\% mIoU on {\it val} set but no improvement over PQ. We do not use this trick for the Mapillary Vistas Challenge. As shown in \tabref{tab:cityscapes_test}, our {\it single} unified Panoptic-DeepLab achieves state-of-the-art results, ranking first at {\it all} three Cityscapes tasks, when comparing with published works. Our model ranks second in the instance segmentation track when also taking into account unpublished entries.

\subsection{Mapillary Vistas}
{\bf Val set:} In \tabref{tab:mapillary_val}, we report Mapillary Vistas {\it val} set results. Our best {\it single} Panoptic-DeepLab model, with multi-scale inputs and left-right flips, outperforms the bottom-up approach, DeeperLab, by 8.3\% PQ, and the top-down approach, Seamless, by 2.6\% PQ. In \tabref{tab:mapillary_backbone}, we report our results with three families of network backbones. We  observe that na\"ive HRNet-W48 slightly under-performs Xception-71. Due to the diverse image resolutions in Mapillary Vistas, we found it important to enrich the context information as well as to keep high-resolution features. Therefore, we propose a simple modification for HRNet \cite{wang2019deep} and Auto-DeepLab \cite{liu2019auto}. For modified HRNet, called HRNet+, we keep its ImageNet-pretrained head and further attach dual-ASPP and dual-decoder modules. For modified Auto-DeepLab, called Auto-DeepLab+, we remove the stride in the original 1/32 branch (which improves PQ by 1\%). To summarize, using Xception-71 strikes the best accuracy and speed trade-off, while HRNet-W48+ achieves the best PQ of 40.6\%. Finally, our ensemble of six models attains a 42.2\% PQ, 18.2\% AP, and 58.7\% mIoU.

{\bf Test set:} \tabref{tab:mapillary_test} summarizes our Mapillary Vistas test set results along with other top-performing methods. Our entry \cite{cheng2019panoptic} with an ensemble of six models attain a performance of 42.7\% PQ, outperforming the winner of Mapillary Vistas Panoptic Segmentation Challenge in 2018 by 1.5\% PQ. 

\begin{table}[!t]
  \centering
  \scalebox{0.62}{
  \begin{tabular}{c | c | c | c | c | c | c | c }
    \toprule[0.2em]
    Method & Flip & MS  & PQ (\%) & $\text{PQ}^{\text{Th}}$ (\%) & $\text{PQ}^{\text{St}}$ (\%) & AP (\%) & mIoU (\%)\\
    \toprule[0.2em]
    TASCNet~\cite{li2018learning} & & & 32.6 & 31.1 & 34.4 & 18.5 & -\\
    TASCNet~\cite{li2018learning} & \cmark & \cmark & 34.3 & \textbf{34.8} & 33.6 & \textbf{20.4} & -\\
    AdaptIS~\cite{sofiiuk2019adaptis} & \cmark & & 35.9 & 31.5 & 41.9 & - & -\\
    Seamless~\cite{porzi2019seamless} & & & 37.7 & 33.8 & 42.9 & 16.4 & 50.4\\
    \midrule
    DeeperLab~\cite{yang2019deeperlab} & & & 32.0 & - & - & - & 55.3\\
    \midrule\midrule
    Panoptic-DeepLab & & & 37.7 & 30.4 & 47.4 & 14.9 & 55.4 \\
    Panoptic-DeepLab & \cmark & & 38.0 & 30.6 & 47.9 & 15.2 & 55.8 \\
    Panoptic-DeepLab & \cmark & \cmark & \textbf{40.3} & 33.5 & \textbf{49.3} & 17.2 & \textbf{56.8} \\
    \bottomrule[0.1em]
  \end{tabular}
  }
  \caption{Mapillary Vistas {\it val} set. {\bf Flip:} Adding left-right flipped inputs. {\bf MS:} Multiscale inputs.}
  \label{tab:mapillary_val}
  \vspace{-2mm}
\end{table}

\begin{table}[!t]
  \centering
  \scalebox{0.62}{
  \begin{tabular}{c | c | c | c | c | c }
    \toprule[0.2em]
    Backbone & Params (M) & M-Adds (B) & PQ (\%) & AP (\%) & mIoU (\%) \\
    \toprule[0.2em]
    Xception-65 & 44.31 & 1054.05 & 39.2 & 16.4 & 56.9 \\
    Xception-71 & 46.73 & 1264.32 & 40.3 & 17.2 & 56.8 \\
    \midrule
    HRNet-W48~\cite{wang2019deep} & 71.66 & 2304.87 & 39.3 & 17.2 & 55.4 \\
    HRNet-W48+ & 88.87 & 2208.04 & 40.6 & 17.8 & 57.6 \\
    HRNet-W48+ (Atrous) & 88.87 & 2972.02 & 40.5 & 17.7 & 57.4\\
    HRNet-Wider+ & 60.05 & 1315.70 & 40.0 & 17.0 & 57.0 \\
    HRNet-Wider+ (Atrous) & 60.05 & 1711.69 & 39.7 & 16.8 & 56.5 \\
    \midrule
    Auto-DeepLab-L+ & 41.54 & 1493.78 & 39.3 & 15.8 & 56.9 \\
    Auto-DeepLab-XL+ & 71.98 & 2378.17 & 40.3 & 16.3 & 57.1 \\
    Auto-DeepLab-XL++ & 72.16 & 2386.81 & 40.3 & 16.9 & 57.6 \\
    \midrule\midrule
    Ensemble (top-6 models) & - & - & 42.2 & 18.2 & 58.7 \\
    \bottomrule[0.1em]
  \end{tabular}
  }
  \caption{Mapillary Vistas {\it val} set with different backbones. {\bf HRNet-W48+:} Modified HRNet-W48 with ImageNet-pretraining head kept. {\bf HRNet-W48+ (Atrous):} Additionally apply atrous convolution with rate 2 in the output stride 32 branch of HRNet. {\bf HRNet-Wider+:} A wider version of HRNet using separable convolution with large channels. The ImageNet-pretraining head is also kept. {\bf HRNet-Wider+ (Atrous):} Additionally apply atrous convolution with rate 2 in the output stride 32 branch. {\bf Auto-DeepLab-L+:} Auto-DeepLab with $F=48$ and remove the stride in the original output stride 32 path. {\bf Auto-DeepLab-XL+:} Auto-DeepLab with $F=64$ and remove the stride in the original output stride 32 path. {\bf Auto-DeepLab-XL++:} Additionally exploit low-level features from output stride 8 endpoint in the decoder module. We employ dual-ASPP and dual-decoder modules for all model variants except {\bf HRNet-W48} which follows the original design in \cite{wang2019deep}. Results are obtained with multi-scale and left-right flipped inputs. M-Adds are measured \wrt a $2177\times2177$ input.}
  \label{tab:mapillary_backbone}
  \vspace{-2mm}
\end{table}

\begin{table}[!t]
  \centering
  \scalebox{0.60}{
  \begin{tabular}{c | c c c | c c c | c c c}
    \toprule[0.2em]
    Method & PQ & SQ & RQ & PQ$^{\text{Th}}$ & SQ$^{\text{Th}}$ & RQ$^{\text{Th}}$ & PQ$^{\text{St}}$ & SQ$^{\text{St}}$ & RQ$^{\text{St}}$ \\
    \toprule[0.2em]
    DeeperLab \cite{yang2019deeperlab} & 31.6 & 75.5 & 40.1 & 25.0 & 73.4 & 33.1 & 40.3 & 78.3 & 49.3 \\
    AdaptIS~\cite{sofiiuk2019adaptis} & 36.8 & 76.0 & 46.3 & 33.3 &75.2 & 42.6 & 41.4 & 77.1 & 51.3\\
    \midrule\midrule
    TRI-ML (2018$:2^{nd}$) & 38.7 & 78.1 & 48.4 & 39.0 & 79.7 & 48.9 & 38.2 & 75.9 & 47.9\\
    Team R4D (2018$:1^{st}$) & 41.2 & 79.1 & 50.8 & 37.9 & 79.7 & 47.1 & 45.6 & 78.4 & 55.8\\
    \midrule
    Panoptic-DeepLab & 42.7 & 78.1 & 52.5 & 35.9 & 75.3 & 46.0 & 51.6 & 81.9 & 61.2\\
    \bottomrule[0.1em]
  \end{tabular}
  }
  \caption{Performance on Mapillary Vistas {\it test} set.}
  \label{tab:mapillary_test}
  \vspace{-2mm}
\end{table}

\subsection{COCO}
{\bf Val set:} In \tabref{tab:coco_val}, we report COCO {\it val} set result. With a single scale inference, our Panoptic-DeepLab outperforms the previous best bottom-up SSAP by $3.2\%$ PQ and DeeperLab \cite{yang2019deeperlab} by 5.9\% PQ. With multi-scale inference and horizontal flip, Panoptic-DeepLab achieves 41.2\% PQ, setting a new state-of-the-art performance for bottom-up methods, and performing comparably with top-down methods.

{\bf Test-dev set:} In \tabref{tab:coco_test}, we report COCO {\it test-dev} set result. Our Panoptic-DeepLab is $4.5\%$ PQ better than the previous best bottom-up SSAP on COCO and our $41.4\%$ PQ is comparable to most top-down methods without using heavier backbone~\cite{xie2017aggregated} or deformable convolution~\cite{dai2017deformable}.

\begin{table}[!t]
  \centering
  \scalebox{0.62}{
  \begin{tabular}{c | c | c | c | c | c | c }
    \toprule[0.2em]
    Method & Backbone & Flip & MS  & PQ (\%) & $\text{PQ}^{\text{Th}}$ (\%) & $\text{PQ}^{\text{St}}$ (\%)\\
    \toprule[0.2em]
    AUNet~\cite{li2018attention} & ResNet-50 \cite{he2016deep} & & & 39.6 & 49.1 & 25.2\\
    Panoptic-FPN~\cite{kirillov2019panoptic} & ResNet-101 & & & 40.3 & 47.5 & 29.5\\
    AdaptIS~\cite{sofiiuk2019adaptis} & ResNeXt-101 \cite{xie2017aggregated} & \cmark & & 42.3 & 49.2 & 31.8\\
    UPSNet~\cite{xiong2019upsnet} & ResNet-50 & & & 42.5 & 48.5 & 33.4 \\
    Detectron2~\cite{wu2019detectron2} & ResNet-101 & & & 43.0 & - & - \\
    UPSNet~\cite{xiong2019upsnet} & ResNet-50 & \cmark & \cmark & 43.2 & 49.1 & 34.1 \\
    \midrule
    DeeperLab~\cite{yang2019deeperlab} & Xception-71 & & & 33.8 & - & - \\
    SSAP~\cite{gao2019ssap} & ResNet-101 & \cmark & \cmark & 36.5 & - & - \\
    \midrule\midrule
    Panoptic-DeepLab & Xception-71 & & & 39.7 & 43.9 & 33.2 \\
    Panoptic-DeepLab & Xception-71 & \cmark & & 40.2 & 44.4 & 33.8 \\
    Panoptic-DeepLab & Xception-71 & \cmark & \cmark & 41.2 & 44.9 & 35.7 \\
    \bottomrule[0.1em]
  \end{tabular}
  }
  \caption{COCO {\it val} set. {\bf Flip:} Adding left-right flipped inputs. {\bf MS:} Multiscale inputs.}
  \label{tab:coco_val}
  \vspace{-2mm}
\end{table}

\begin{table}[!t]
  \centering
  \scalebox{0.66}{
  \begin{tabular}{c | c | c | c | c | c | c }
    \toprule[0.2em]
    Method & Backbone & Flip & MS  & PQ (\%) & $\text{PQ}^{\text{Th}}$ (\%) & $\text{PQ}^{\text{St}}$ (\%) \\
    \toprule[0.2em]
    TASCNet~\cite{li2018learning} & ResNet-50 & & & 40.7 & 47.0 & 31.0 \\
    Panoptic-FPN~\cite{kirillov2019panoptic} & ResNet-101 & & & 40.9 & 48.3 & 29.7  \\
    AdaptIS~\cite{sofiiuk2019adaptis} & ResNeXt-101 & \cmark & & 42.8 & 53.2 & 36.7 \\
    AUNet~\cite{li2018attention} & ResNeXt-152 & & & 46.5 & 55.8 & 32.5 \\
    UPSNet~\cite{xiong2019upsnet} & DCN-101 \cite{dai2017deformable} & \cmark & \cmark & 46.6 & 53.2 & 36.7 \\
    \midrule
    DeeperLab~\cite{yang2019deeperlab} & Xception-71 & & & 34.3 & 37.5 & 29.6 \\
    SSAP~\cite{gao2019ssap} & ResNet-101 & \cmark & \cmark & 36.9 & 40.1 & 32.0 \\
    \midrule\midrule
    Panoptic-DeepLab & Xception-71 & \cmark & \cmark & 41.4 & 45.1 & 35.9 \\
    \bottomrule[0.1em]
  \end{tabular}
  }
  \caption{COCO {\it test-dev} set. {\bf Flip:} Adding left-right flipped inputs. {\bf MS:} Multiscale inputs.}
  \label{tab:coco_test}
  \vspace{-2mm}
\end{table}

\subsection{Runtime}
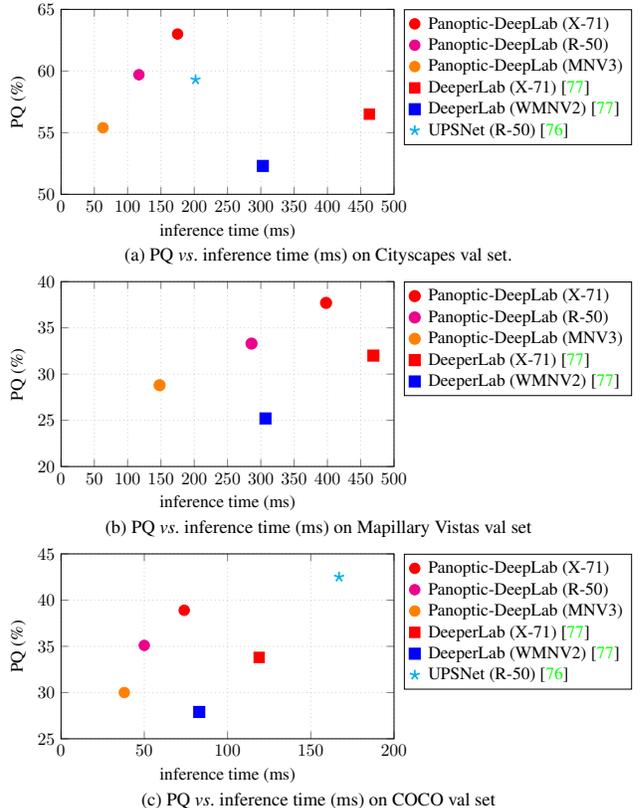
\begin{figure}
    \hspace{-4mm}
    \begin{tabular} {c}
    \vspace{-2mm}
    \resizebox{\linewidth}{!}{\begin{tikzpicture}[/pgfplots/width=1\linewidth, /pgfplots/height=0.64\linewidth, /pgfplots/legend pos=south east]
    \begin{axis}[
        ymin=50,ymax=65,xmin=0,xmax=500,
        xlabel=inference time (ms),
        ylabel=PQ (\%),
		font=\normalsize,
        grid=both,
		grid style=dotted,
        legend columns=1,
	    xtick={0,50,...,500},
	    ytick={0,5,...,70},
        legend pos= outer north east,
        legend cell align={left}
        ]

	    \addplot[red,mark=*,only marks,mark size=3] coordinates{(175, 63)};
        \addlegendentry{\hphantom{i}Panoptic-DeepLab (X-71)}
        
        \addplot[magenta,mark=*,only marks,mark size=3] coordinates{(117, 59.7)};
        \addlegendentry{\hphantom{i}Panoptic-DeepLab (R-50)}

        \addplot[orange,mark=*,only marks,mark size=3] coordinates{(63, 55.4)};
        \addlegendentry{\hphantom{i}Panoptic-DeepLab (MNV3)}

        \addplot[red,mark=square*,only marks,mark size=3] coordinates{(463, 56.5)};
        \addlegendentry{\hphantom{i}DeeperLab (X-71) \cite{yang2019deeperlab}}
       
        \addplot[blue,mark=square*,only marks,line width=1, mark size=3] coordinates{(303, 52.3)};
        \addlegendentry{\hphantom{i}DeeperLab (WMNV2) \cite{yang2019deeperlab}}

        \addplot[cyan,mark=star, mark size=3,only marks, line width=1] coordinates{(202, 59.3)};
        \addlegendentry{\hphantom{i}UPSNet (R-50) \cite{xiong2019upsnet}}
        
    \end{axis}
\end{tikzpicture}} \\
    {\scriptsize (a) PQ \vs inference time (ms) on Cityscapes val set.} \\
    \vspace{-2mm}
    \resizebox{\linewidth}{!}{\begin{tikzpicture}[/pgfplots/width=1\linewidth, /pgfplots/height=0.64\linewidth, /pgfplots/legend pos=south east]
    \begin{axis}[
        ymin=20,ymax=40,xmin=0,xmax=500,
        xlabel=inference time (ms),
        ylabel=PQ (\%),
		font=\normalsize,
        grid=both,
		grid style=dotted,
        legend columns=1,
	    xtick={0,50,...,500},
	    ytick={0,5,...,50},
        legend pos= outer north east,
        legend cell align={left}
        ]
		
	    \addplot[red,mark=*,mark size=3,only marks,line width=1] coordinates{(398, 37.7)};
        \addlegendentry{\hphantom{i}Panoptic-DeepLab (X-71)}
        
        \addplot[magenta,mark=*,mark size=3,only marks,line width=1] coordinates{(286, 33.3)};
        \addlegendentry{\hphantom{i}Panoptic-DeepLab (R-50)}

        \addplot[orange,mark=*,mark size=3,only marks,line width=1] coordinates{(148, 28.8)};
        \addlegendentry{\hphantom{i}Panoptic-DeepLab (MNV3)}

        \addplot[red,mark=square*,only marks,mark size=3,line width=1] coordinates{(469, 32)};
        \addlegendentry{\hphantom{i}DeeperLab (X-71) \cite{yang2019deeperlab}}
       
        \addplot[blue,mark=square*,mark size=3,only marks,line width=1] coordinates{(307, 25.2)};
        \addlegendentry{\hphantom{i}DeeperLab (WMNV2) \cite{yang2019deeperlab}}
    \end{axis}
\end{tikzpicture}} \\
    {\scriptsize (b) PQ \vs inference time (ms) on Mapillary Vistas val set} \\
    \vspace{-2mm}
    \resizebox{\linewidth}{!}{\begin{tikzpicture}[/pgfplots/width=1\linewidth, /pgfplots/height=0.64\linewidth, /pgfplots/legend pos=south east]
    \begin{axis}[
        ymin=25,ymax=45,xmin=0,xmax=200,
        xlabel=inference time (ms),
        ylabel=PQ (\%),
		font=\normalsize,
        grid=both,
		grid style=dotted,
        legend columns=1,
	    xtick={0,50,...,300},
	    ytick={0,5,...,60},
        legend pos= outer north east,
        legend cell align={left}
        ]

	    \addplot[red,mark=*,only marks,mark size=3] coordinates{(74, 38.9)};
        \addlegendentry{\hphantom{i}Panoptic-DeepLab (X-71)}
    
        \addplot[magenta,mark=*,only marks,mark size=3] coordinates{(50, 35.1)};
        \addlegendentry{\hphantom{i}Panoptic-DeepLab (R-50)}

        \addplot[orange,mark=*,only marks,mark size=3] coordinates{(38, 30.0)};
        \addlegendentry{\hphantom{i}Panoptic-DeepLab (MNV3)}

        \addplot[red,mark=square*,only marks,mark size=3] coordinates{(119, 33.8)};
        \addlegendentry{\hphantom{i}DeeperLab (X-71) \cite{yang2019deeperlab}}
       
        \addplot[blue,mark=square*,only marks,line width=1, mark size=3] coordinates{(83, 27.9)};
        \addlegendentry{\hphantom{i}DeeperLab (WMNV2) \cite{yang2019deeperlab}}

        \addplot[cyan,mark=star, mark size=3,only marks, line width=1] coordinates{(167, 42.5)};
        \addlegendentry{\hphantom{i}UPSNet (R-50) \cite{xiong2019upsnet}}
    \end{axis}
\end{tikzpicture}} \\
    {\scriptsize (c) PQ \vs inference time (ms) on COCO val set} \\
    \end{tabular}
    \caption{PQ \vs Seconds. Our Panoptic-DeepLab model variants attain a better speed/accuracy trade-off across challenging datasets. The inference time is measured {\it end-to-end} from input image to panoptic segmentation output. {\bf X-71:} Xception-71. {\bf R-50:} ResNet-50. {\bf MNV3:} MobileNetV3. Data points from \tabref{tab:runtime}.
    }
    \vspace{-2mm}
    \label{fig:pq_vs_second}
\end{figure}

In \tabref{tab:runtime}, we report the end-to-end runtime (\ie, inference time from an input image to final panoptic segmentation, including {\it all} operations such as merging semantic and instance segmentation) of Panoptic-DeepLab with three different network backbones (MobileNetV3~\cite{howard2019searching}, ResNet-50 \cite{he2016deep}, and Xception-71 \cite{chollet2016xception,dai2017coco}) on all three datasets. The inference speed is measured on a Tesla V100-SXM2 GPU \emph{with batch size of one}. We further plot the speed-accuracy trade-off curve in \figref{fig:pq_vs_second}. Our Panoptic-DeepLab achieves the best trade-off across all three datasets.

\begin{table}[!t]
  \centering
  \scalebox{0.54}{
  \begin{tabular}{c | c | c | c | c | c | c }
    \toprule[0.2em]
    Method & Backbone & Input Size & PQ [val] & PQ [test] & Speed (ms) & M-Adds (B) \\
    \toprule[0.2em]
    \multicolumn{7}{c}{Cityscapes}\\
    \midrule
    DeeperLab~\cite{yang2019deeperlab} & W-MNV2 \cite{sandler2018mobilenetv2} & $1025\times2049$ & 52.3 & - & 303 & - \\
    DeeperLab~\cite{yang2019deeperlab} & Xception-71 & $1025\times2049$ & 56.5 & - & 463 & - \\
    UPSNet~\cite{xiong2019upsnet} & ResNet-50 & $1024\times2048$ & 59.3 & - & 202 & - \\
    \midrule \midrule
    Panoptic-DeepLab & MNV3 & $1025\times2049$ & 55.4 & 54.1 & 63 & 54.17 \\
    Panoptic-DeepLab & ResNet-50 & $1025\times2049$ & 59.7 & 58.0 & 117 & 381.39 \\
    Panoptic-DeepLab & Xception-71 & $1025\times2049$ & 63.0 & 60.7 & 175 & 547.49 \\
    \midrule
    \multicolumn{7}{c}{Mapillary Vistas}\\
    \midrule
    DeeperLab~\cite{yang2019deeperlab} & W-MNV2 & $1441\times1441$ & 25.2 & 25.3 & 307 & - \\
    DeeperLab~\cite{yang2019deeperlab} & Xception-71 & $1441\times1441$ & 32.0 & 31.6 & 469 & - \\
    \midrule \midrule
    Panoptic-DeepLab & MNV3 & $2177\times2177$ & 28.8 & - & 148 & 138.12 \\
    Panoptic-DeepLab & ResNet-50 & $2177\times2177$ & 33.3 & - & 286 & 910.47 \\
    Panoptic-DeepLab & Xception-71 & $2177\times2177$ & 37.7 & - & 398 & 1264.32 \\
    \midrule
    \multicolumn{7}{c}{COCO }\\
    \midrule
    DeeperLab~\cite{yang2019deeperlab} & W-MNV2 & $641\times641$ & 27.9 & 28.1 & 83 & - \\
    DeeperLab~\cite{yang2019deeperlab} & Xception-71 & $641\times641$ & 33.8 & 34.3 & 119 & - \\
    UPSNet~\cite{xiong2019upsnet} & ResNet-50 & $800\times1333$ & 42.5 & - & 167 & - \\
    \midrule \midrule
    Panoptic-DeepLab & MNV3 & $641\times641$ & 30.0 & 29.8 & 38 & 12.24 \\
    Panoptic-DeepLab & ResNet-50 & $641\times641$ & 35.1 & 35.2 & 50 & 77.79 \\
    Panoptic-DeepLab & Xception-71 & $641\times641$ & 38.9 & 38.8 & 74 & 109.21 \\
    Panoptic-DeepLab & Xception-71 & $1025\times1025$ & 39.7 & 39.6 & 132 & 279.25 \\
    \bottomrule[0.1em]
  \end{tabular}
  }
  \caption{End-to-end runtime, including merging semantic and instance segmentation. All results are obtained by (1) a single-scale input without flipping, and (2) built-in TensorFlow library without extra inference optimization. {\bf MNV3:} MobileNet-V3. {\bf PQ [val]:} PQ (\%) on val set. {\bf PQ [test]:} PQ (\%) on test(-dev) set. Note the channels in last block of MNV3 are reduced by a factor of 2  \cite{howard2019searching}.
  }
  \label{tab:runtime}
\end{table}

\begin{figure*}[!t]
\begin{center}
\bgroup 
 \def\arraystretch{0.2} 
 \setlength\tabcolsep{0.2pt}
\begin{tabular}{cccccc}
\includegraphics[width=0.16666666666667\linewidth]{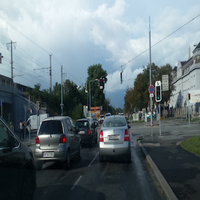} &
\includegraphics[width=0.16666666666667\linewidth]{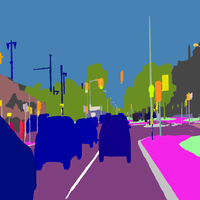} &
\includegraphics[width=0.16666666666667\linewidth]{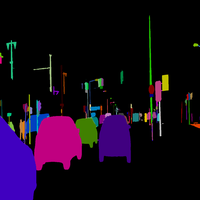} &
\includegraphics[width=0.16666666666667\linewidth]{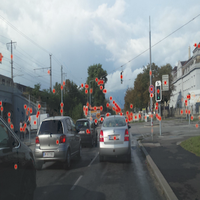} &
\includegraphics[width=0.16666666666667\linewidth]{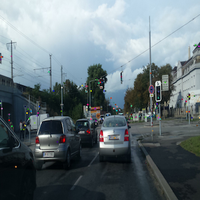} &
\includegraphics[width=0.16666666666667\linewidth]{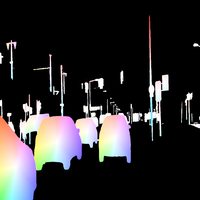} \\
\includegraphics[width=0.16666666666667\linewidth]{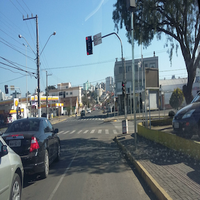} &
\includegraphics[width=0.16666666666667\linewidth]{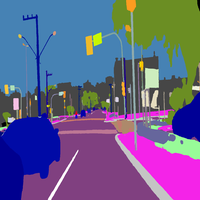} &
\includegraphics[width=0.16666666666667\linewidth]{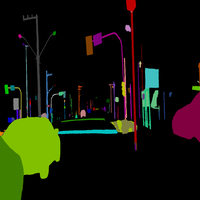} &
\includegraphics[width=0.16666666666667\linewidth]{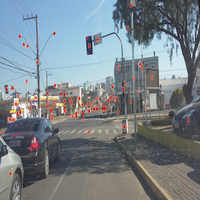} &
\includegraphics[width=0.16666666666667\linewidth]{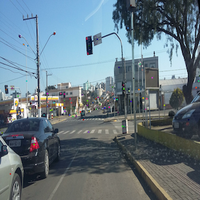} &
\includegraphics[width=0.16666666666667\linewidth]{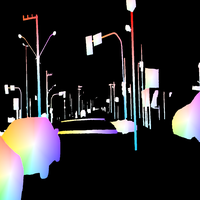} \\
\includegraphics[width=0.16666666666667\linewidth]{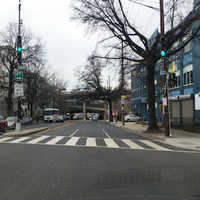} &
\includegraphics[width=0.16666666666667\linewidth]{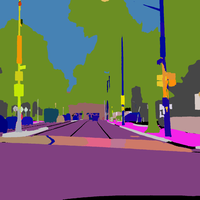} &
\includegraphics[width=0.16666666666667\linewidth]{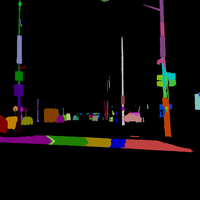} &
\includegraphics[width=0.16666666666667\linewidth]{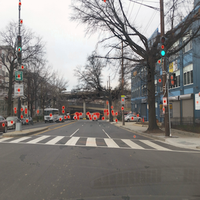} &
\includegraphics[width=0.16666666666667\linewidth]{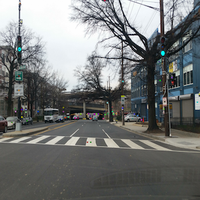} &
\includegraphics[width=0.16666666666667\linewidth]{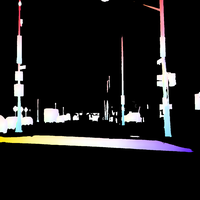} \\
\includegraphics[width=0.16666666666667\linewidth]{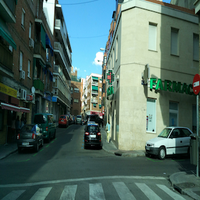} &
\includegraphics[width=0.16666666666667\linewidth]{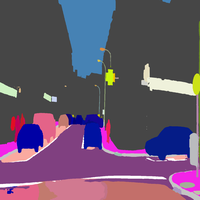} &
\includegraphics[width=0.16666666666667\linewidth]{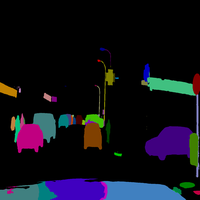} &
\includegraphics[width=0.16666666666667\linewidth]{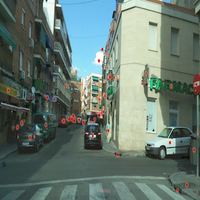} &
\includegraphics[width=0.16666666666667\linewidth]{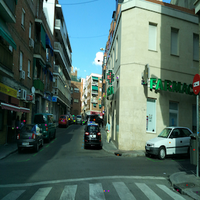} &
\includegraphics[width=0.16666666666667\linewidth]{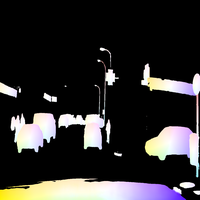} \\
Image & Panoptic Prediction & Instance Prediction & Heatmap Prediction & Center Prediction & Center Regression \\ 
\end{tabular} \egroup 
\end{center}
\caption{Visualization of Panoptic-DeepLab with Xception-71 on Mapillary Vistas {\it val} set. Only single scale inference is used and the model achieves $37.7\%$ PQ. We encode 2D offset vectors into RGB values, same as~\cite{baker2011database}. The cross road in last 2 rows is segmented into multiple instances due to large scale variation. 
More visualizations are included in Appendix~\ref{sec:more_vis}.
}
\label{fig:failure}
\vspace{-2mm}
\end{figure*}

\subsection{Discussion}
Herein, we list a few interesting aspects in the hope of inspiring future works on bottom-up panoptic segmentation.

{\bf Scale variation:} 
\figref{fig:failure} shows visualization of Panoptic-DeepLab. In particular, the cross road (in last 2 rows), with a large scale variation, is segmented  into multiple small instances. On the other hand, top-down methods handle scale variation to some extent by the ROIPooling \cite{girshick2015fast} or ROIAlign \cite{he2017mask} operations which normalize regional features to a {\it canonical} scale \cite{girshick2014rich,ren2015faster}. Additionally, incorporating scale-aware information to feature pyramid \cite{lin2017feature} or image pyramid \cite{analysissnip2017} may improve the performance of bottom-up methods.

{\bf PQ$^{\text{Thing}}$ \vs PQ$^{\text{Stuff}}$:} As shown in \tabref{tab:mapillary_test} and \tabref{tab:coco_test}, Panoptic-DeepLab has higher PQ$^{\text{Stuff}}$ but lower PQ$^{\text{Thing}}$ when compared with other top-down approaches which better handle instances of large scale variation as discussed above. Combining the best from both bottom-up and top-down approaches is thus interesting to explore but beyond the scope of current work.

{\bf Panoptic \vs instance annotations:} Most bottom-up panoptic segmentation methods only exploit the panoptic annotations. We notice there are two types of annotations in the COCO dataset, panoptic annotations and instance annotations. The former do not allow overlapping masks (thus creating occlusions among masks), while the latter allows overlaps, which might make the training target easier to optimize, similar to amodal segmentation \cite{zhu2017semantic,li2016amodal}. 

{\bf End-to-end training:} Current bottom-up panoptic segmentation methods still require some post-processing steps to obtain the final panoptic segmentation, which may make it hard to end-to-end train the whole system.

\vspace{-2mm}
\section{Conclusion}
We have presented Panoptic-DeepLab, a simple, strong, and fast baseline for bottom-up panoptic segmentation. Panoptic-DeepLab is simple in design, requiring only three loss functions during training and adds marginal parameters to a modern semantic segmentation model. Panoptic-DeepLab is the first bottom-up and single-shot panoptic segmentation model that attains state-of-the-art performance on several public benchmarks, and delivers near real-time end-to-end inference speed. We hope our simple and effective model could establish a solid baseline and further benefit the research community.

\appendix
\section{HRNet Variant}
We introduce our modifications to the HRNet~\cite{sun2019deep,wang2019deep} that are used in our ensemble model \cite{cheng2019panoptic} for Mapillary Vistas. All hyper-parameters for training HRNet variants are the same as Xception, except that the learning rate is set to $7.5e-4$.

\subsection{HRNet}
\begin{figure}[t!]
    \centering
    \includegraphics[width=0.3\textwidth]{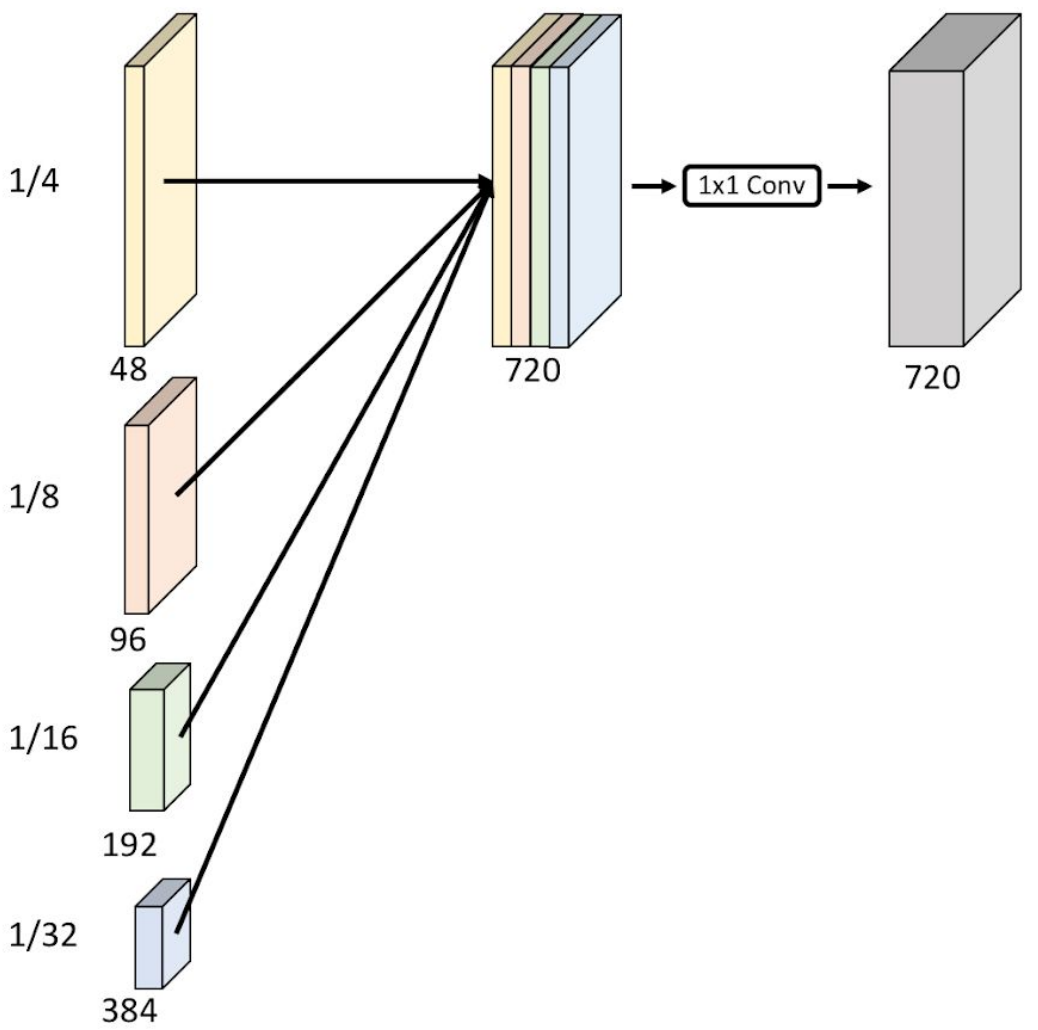}
    \caption{Semantic segmentation head proposed in HRNet~\cite{wang2019deep}.}
    \label{fig:hrnet_segmentation}
\end{figure}

\begin{figure}[t!]
	\centering
	\begin{minipage}[b]{0.99\linewidth}
    \includegraphics[width=1.0\linewidth]{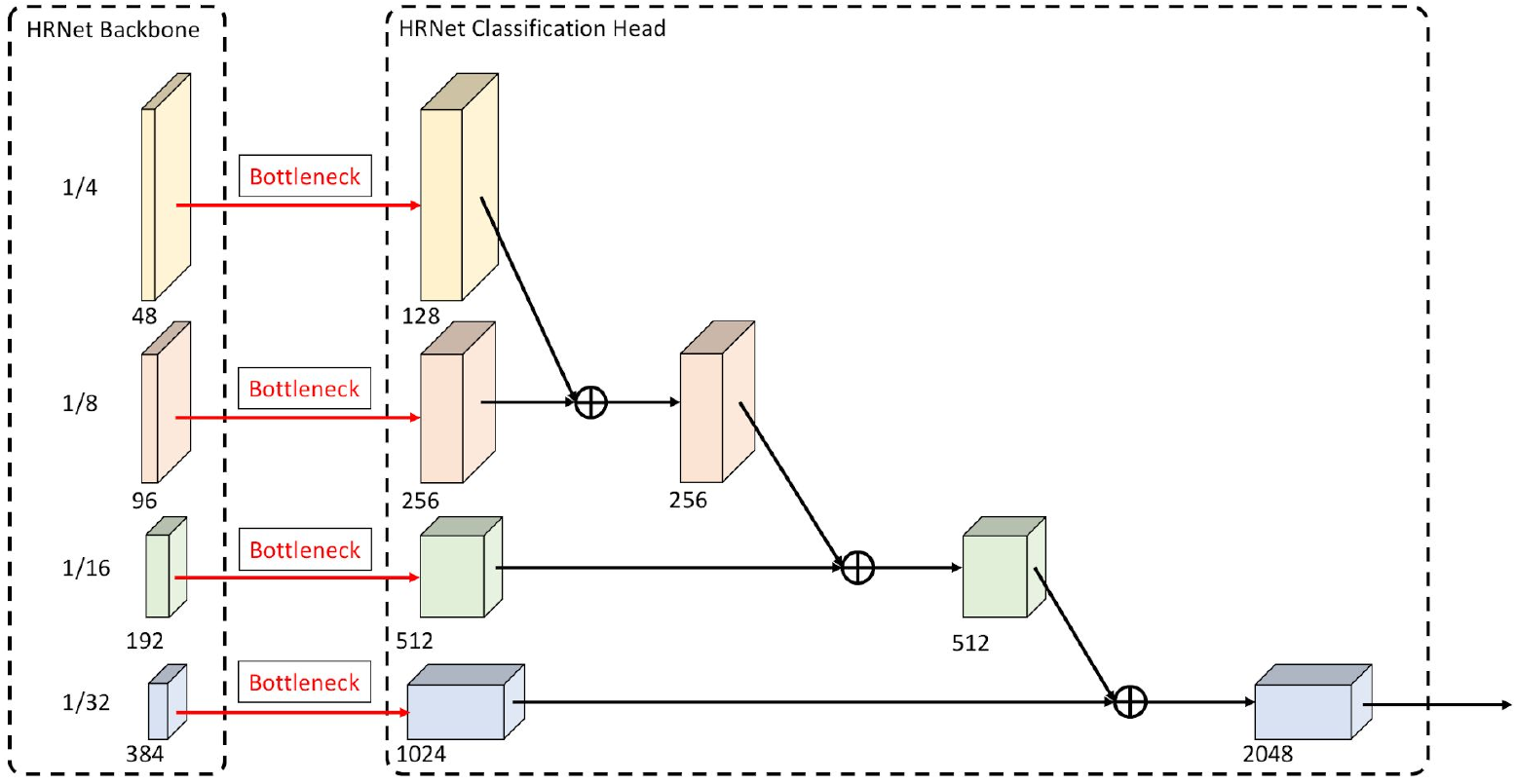}
    \\
    {\footnotesize(a) Image classification head proposed in HRNet~\cite{wang2019deep}, which is discarded after pre-training on ImageNet.}
    \\
    \end{minipage}
    \begin{minipage}[b]{0.99\linewidth}
    \includegraphics[width=1.0\linewidth]{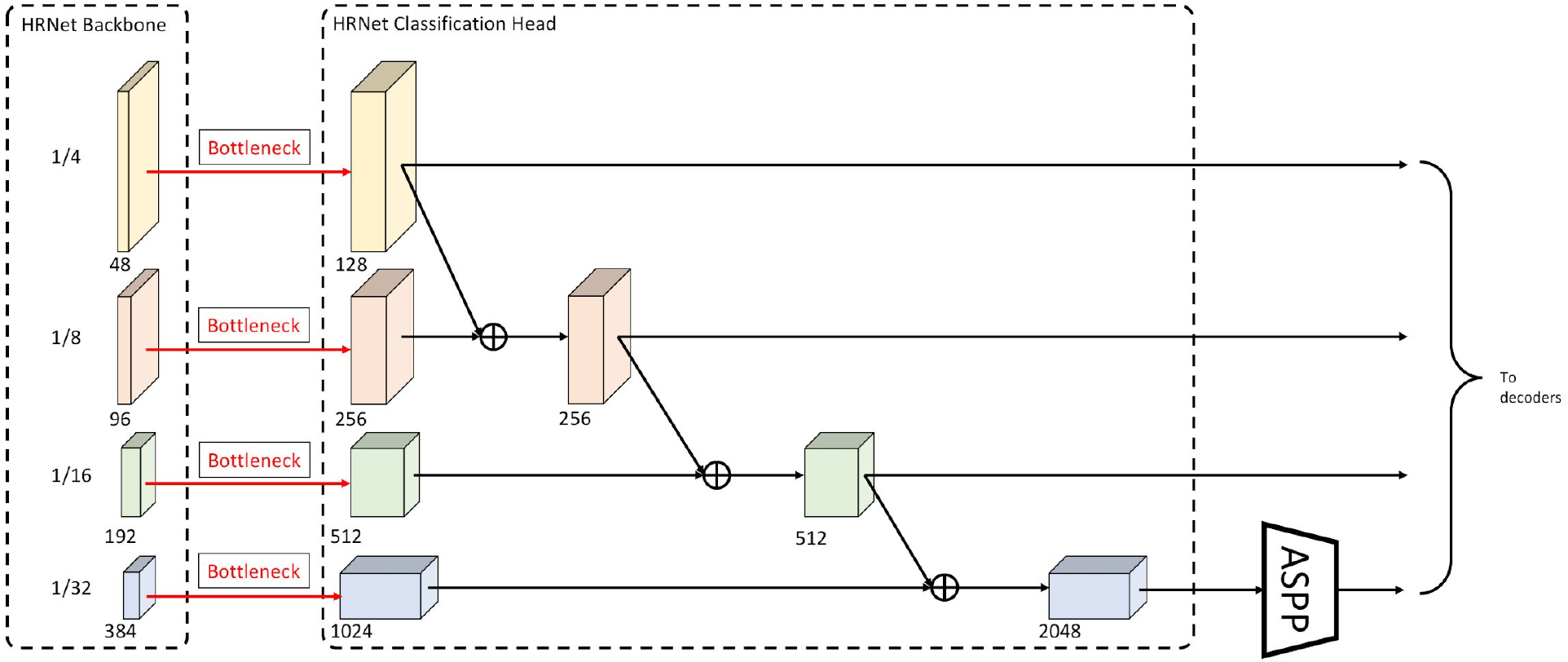}
    \\
    {\footnotesize(b) Our proposed HRNet+, which keeps the image classification head and attaches the ASPP module as well as the decoder module for segmentation tasks.}
    \\
    \end{minipage}
    \begin{minipage}[b]{0.99\linewidth}
    \includegraphics[width=1.0\linewidth]{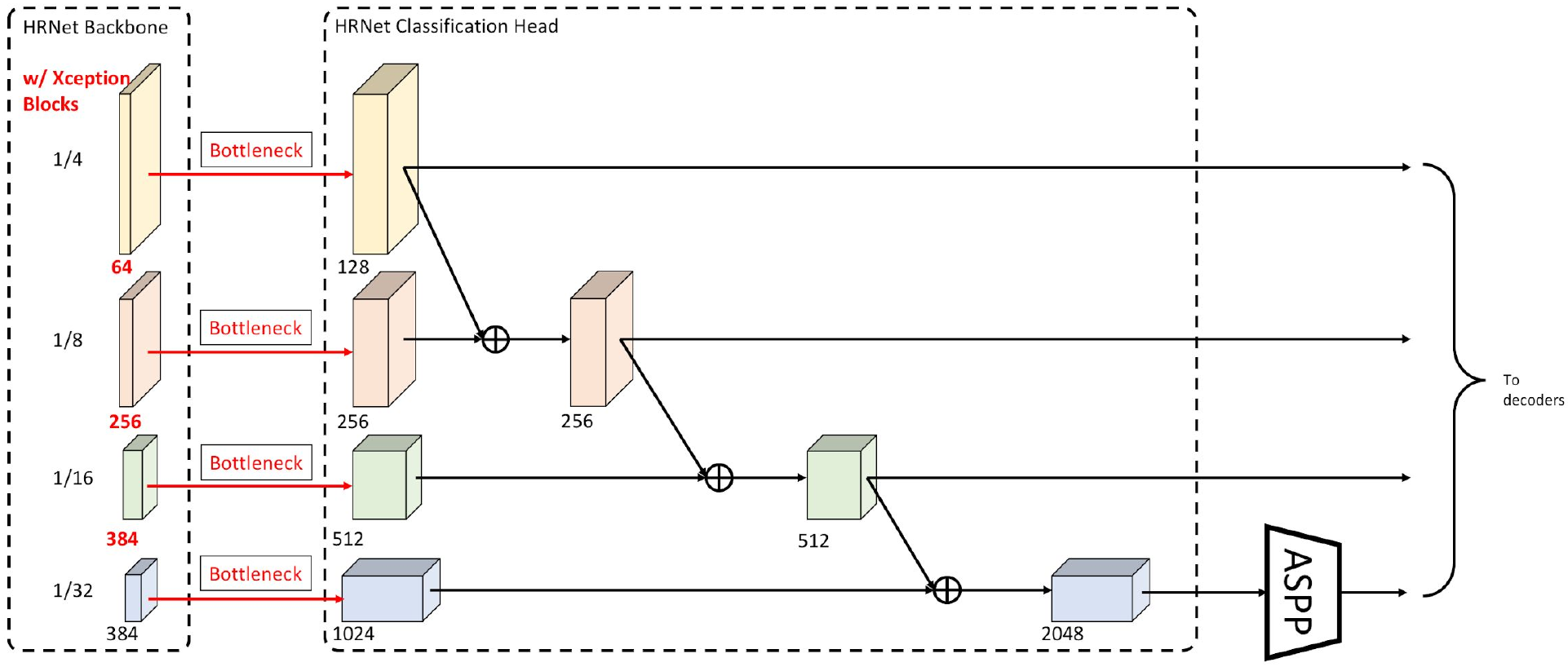}
    \\
    {\footnotesize(c) Our proposed HRNet-Wider+, which reduces the model parameters and computations by adopting the Xception module.}
    \\
    \end{minipage}
	\caption{Demonstration of our proposed variants of HRNet~\cite{wang2019deep}.}
	\label{fig:hrnet_variants}
\end{figure}

The original segmentation head for HRNet is shown in \figref{fig:hrnet_segmentation}. Features from all four resolutions are first upsampled to the 1/4 resolution and concatenated, followed by another $1\times1$ convolution to fuse features.

To pre-train the HRNet on ImageNet~\cite{deng2009imagenet}, Wang \etal~\cite{wang2019deep} designed a specific image classification head which gradually downsamples the feature maps, as shown in \figref{fig:hrnet_variants}~(a). Specifically, a bottleneck residual module \cite{he2016deep} is applied to every output resolution to increase the channels. The feature map from the finest spatial resolution (\ie, 1/4 resolution) is then downsampled by sequentially using a $3\times3$ convolution with stride 2. At the final 1/32 resolution feature map, a global average pooling and a fully connected layer are attached for ImageNet classification. 

\begin{table*}[th!]
  \centering
  \scalebox{0.90}{
  \begin{tabular}{c | c | c | c | c | c | c | c | c }
    \toprule[0.2em]
    Decoder & Backbone & Input Size & PQ ($\%$) & AP ($\%$) & mIoU ($\%$) & Speed (ms) & Params (M) & M-Adds (B) \\
    \toprule[0.2em]
    DeepLabV3+~\cite{deeplabv3plus2018} & Xception-71 & $1025\times2049$ & 62.5 & 34.5 & 80.2 & 176 & 46.61 & 553.41\\
    \midrule
    Panoptic-DeepLab & Xception-71 & $1025\times2049$ & 63.0 & 35.3 & 80.5 & 175 & 46.72 & 547.49 \\
    \bottomrule[0.1em]
  \end{tabular}
  }
  \caption{Comparison between the decoder design of DeepLabV3+~\cite{deeplabv3plus2018} and Panoptic-DeepLab on Cityscapes validation set.
  }
  \label{tab:deeplabv3plus}
\end{table*}

\subsection{HRNet+}
After pre-training on ImageNet, Wang \etal~\cite{wang2019deep} removed the image classification head. However, we observe that the classification head takes around $20\%$ of the total parameters, which is a waste of information if discarded. Therefore, we propose to keep this classification head in our modified HRNet+ (\figref{fig:hrnet_variants}~(b)). Starting from the image classification HRNet, we replace the final global average pooling and linear classifier with an ASPP module, and build a similar decoder as shown in 
\figref{fig:network_architecture} 
of main paper with some differences that the output stride of encoder is now 32 instead of 16 and we introduce one more encoder feature map of stride 16 to the decoder by first projecting its channels to 96.

\subsection{HRNet-Wider+}
We additionally propose HRNet-Wider+ (\figref{fig:hrnet_variants}~(c)) that replaces the basic residual module \cite{he2016deep} with the Xception module \cite{chollet2016xception}, significantly reducing the model parameters and computation FLOPs at the cost of marginal degradation in performance. Additionally, we employ the number of channels $\{64, 256, 384, 384\}$ for each resolution (instead of $\{48, 96, 192, 384\}$).

\subsection{Atrous HRNet}
Another modification of HRNet that we have explored is referred to as HRNet+ (Atrous), where we remove all the downsampling operations that generate 1/32 resolution feature maps and apply atrous convolution with rate equal to 2 in that branch. This modification increases the computation FLOPs but does not improve the performance compared to its HRNet+ counterpart.

\begin{figure}[t!]
    \centering
    \includegraphics[width=0.48\textwidth]{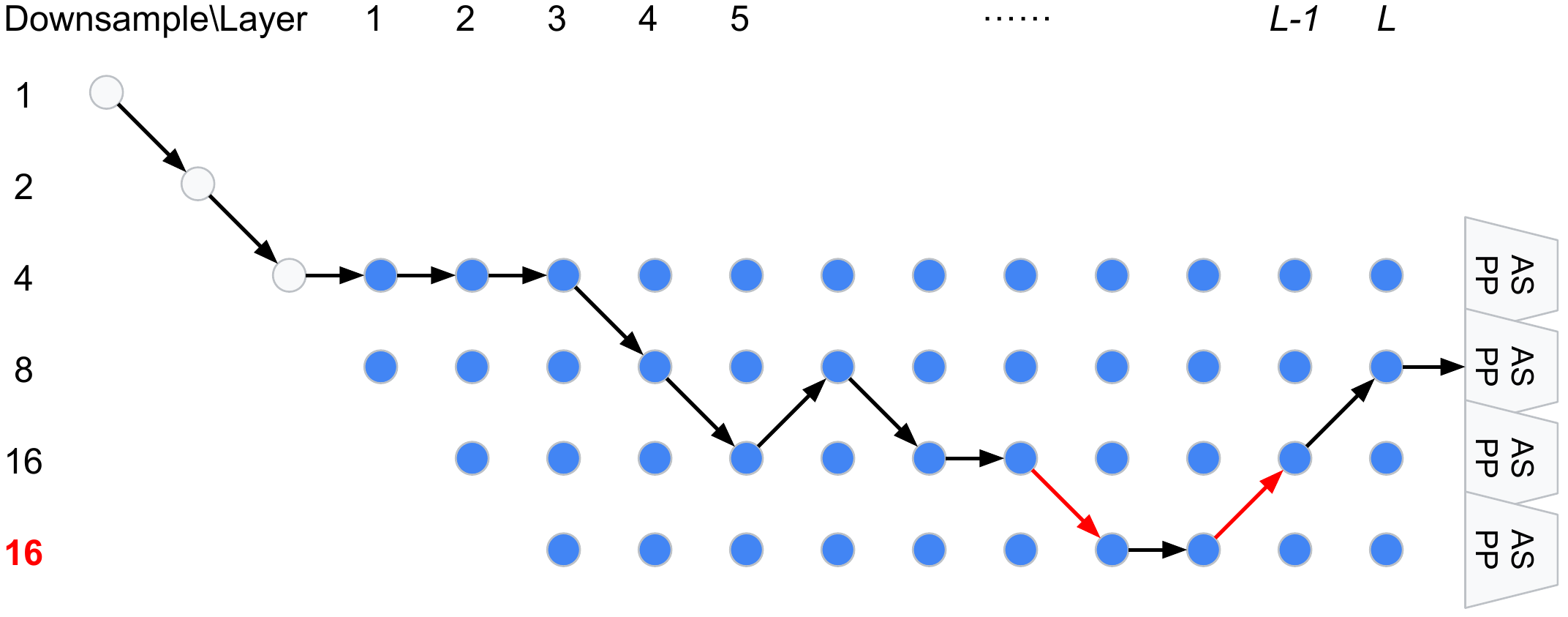}
    \caption{Our proposed Auto-DeepLab+, which keeps the high spatial resolution of feature maps by removing the last stride, \ie, no spatial resolution changes marked in the red arrows.}
    \label{fig:autodeeplab_plus}
\end{figure}

\section{Auto-DeepLab Variant}
We make a simple modification to the Auto-DeepLab~\cite{liu2019auto} in \figref{fig:autodeeplab_plus} by removing the stride in the convolution that generates the 1/32 feature map in order to keep high spatial resolution within the network backbone. We find this modification improves $1\%$ PQ on Mapillary Vistas validation set.

\section{Comparison with DeepLabV3+ decoder}
As mentioned in the main paper that the decoder of Panoptic-DeepLab is slightly different from the one in DeepLabv3+~\cite{deeplabv3plus2018}. Herein, we compare their performance on Cityscapes validation set, as shown in \tabref{tab:deeplabv3plus}. Panoptic-DeepLab outperforms DeepLabv3+ by $0.5\%$ PQ, $0.8\%$ AP, and $0.3\%$ mIOU, showing more improvement in the instance segmentation task. Additionally, Panoptic-DeepLab is slightly faster than DeepLabv3+ at the cost of extra marginal parameters.

\section{Comparison with different instance scores}
\begin{table}[!ht]
  \centering
  \scalebox{0.80}{
  \begin{tabular}{c | c | c | c }
    \toprule[0.2em]
    Instance score & PQ (\%) & AP (\%) & mIoU (\%)\\
    \toprule[0.2em]
    Score(Objectness) & 63.0 & 28.9 & 80.5 \\
    Score(Class) & 63.0 & 35.1 & 80.5 \\
    Score(Objectness) x Score(Class) & 63.0 & \textbf{35.3} & 80.5 \\
    \bottomrule[0.1em]
  \end{tabular}
  }
  \caption{Ablation study on using different confidence scores. Note the choice of confidence scores only affects AP.}
  \label{tab:score}
\end{table}

In \tabref{tab:score}, we experiment with different confidence scores when evaluating instance segmentation results. We found that using Score(Objectness) alone leads to 28.9\% AP, Score(Class) alone produces 35.1\% AP, while employing Score(Objectness) $\times$ Score(Class) generates the best result (35.3\% AP). We would like to highlight that the choice of different confidence score does not affect our final mIoU and PQ results, since our Panoptic-DeepLab does not produce overlapping predictions and therefore does not require a confidence score to rank predictions (or to resolve the conflict among overlapping predictions) like Panoptic-FPN~\cite{kirillov2019panoptic}. Confidence score is only used in computing AP to rank instance mask predictions. Since it is only used for the purpose of ranking, the confidence score does not necessarily need to be a probability.

\section{Instance and Panoptic Annotation}
\begin{figure}[t!]
    \centering
    \includegraphics[width=0.45\textwidth]{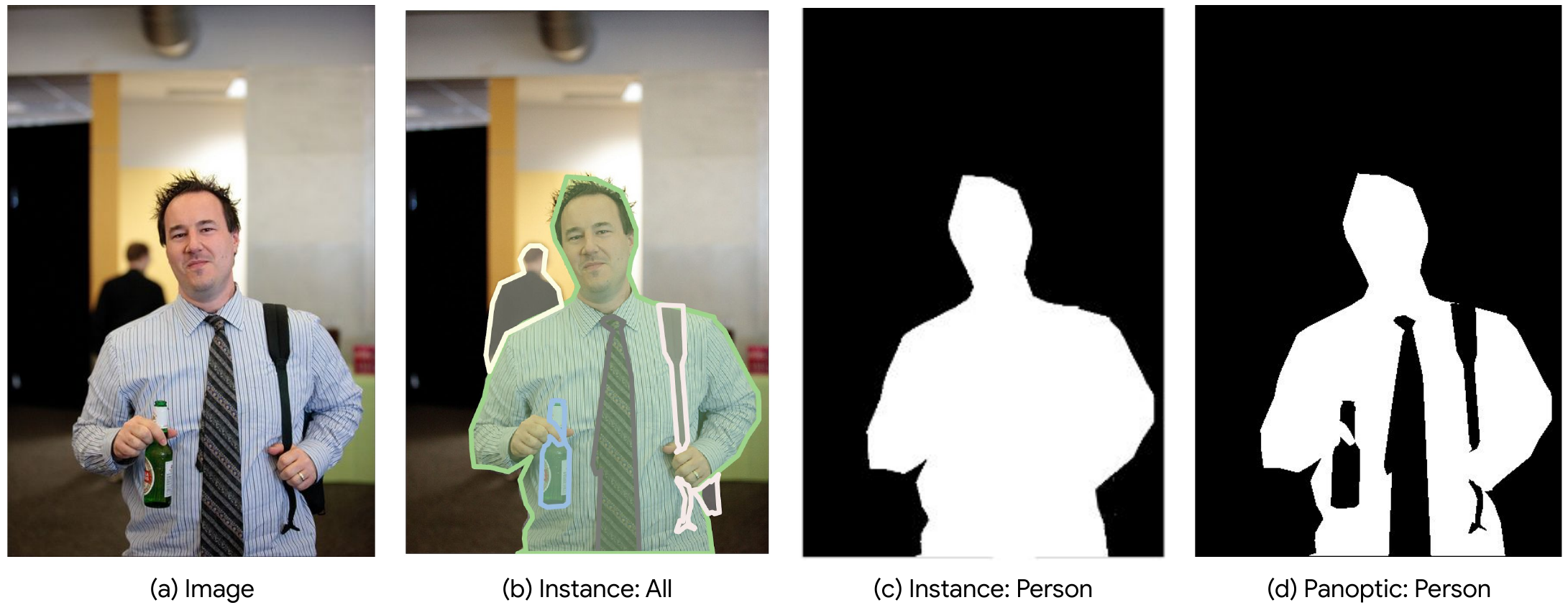}
    \caption{Illustration of the difference between instance and panoptic annotation on COCO.}
    \label{fig:annotation}
\end{figure}

\figref{fig:annotation} shows an example to illustrate the difference between instance annotation and panoptic annotation on the COCO dataset. Instance annotation, unlike panoptic annotation, allows overlapping groundtruth masks. For example, the `person' mask ignores the existence of the `tie' and `bottle' masks in the instance annotation, while the `person' mask has occlusions caused by other instances in the panoptic annotation.

We notice that all top-down methods based on Mask R-CNN~\cite{he2017mask} use the \emph{instance annotation} ~\cite{li2018attention,kirillov2019panoptic,xiong2019upsnet} when trained on COCO, while bottom-up methods~\cite{yang2019deeperlab} including our Panoptic-DeepLab use the \emph{panoptic annotation} on all datasets.

\section{More Visualization}
\label{sec:more_vis}
\begin{figure*}[!t]
\begin{center}
\bgroup 
 \def\arraystretch{0.2} 
 \setlength\tabcolsep{0.2pt}
\begin{tabular}{cccc}
\includegraphics[width=0.25\linewidth]{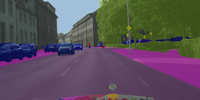} &
\includegraphics[width=0.25\linewidth]{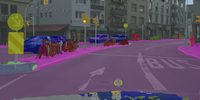} &
\includegraphics[width=0.25\linewidth]{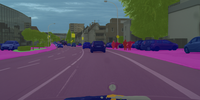} &
\includegraphics[width=0.25\linewidth]{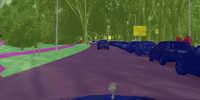}\\
\includegraphics[width=0.25\linewidth]{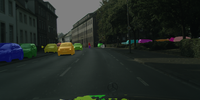} &
\includegraphics[width=0.25\linewidth]{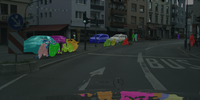} &
\includegraphics[width=0.25\linewidth]{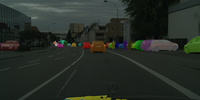} &
\includegraphics[width=0.25\linewidth]{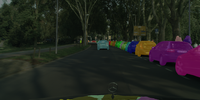}\\
\includegraphics[width=0.25\linewidth]{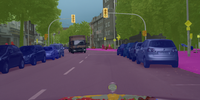} &
\includegraphics[width=0.25\linewidth]{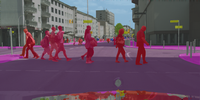} &
\includegraphics[width=0.25\linewidth]{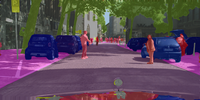} &
\includegraphics[width=0.25\linewidth]{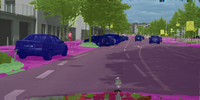}\\
\includegraphics[width=0.25\linewidth]{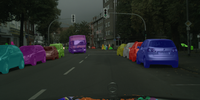} &
\includegraphics[width=0.25\linewidth]{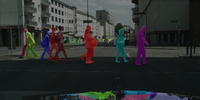} &
\includegraphics[width=0.25\linewidth]{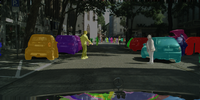} &
\includegraphics[width=0.25\linewidth]{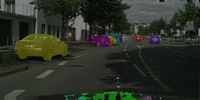}\\
\includegraphics[width=0.25\linewidth]{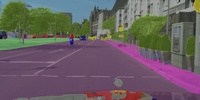} &
\includegraphics[width=0.25\linewidth]{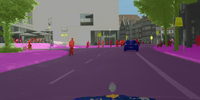} &
\includegraphics[width=0.25\linewidth]{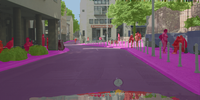} &
\includegraphics[width=0.25\linewidth]{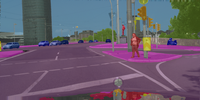}\\
\includegraphics[width=0.25\linewidth]{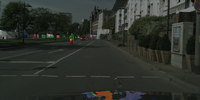} &
\includegraphics[width=0.25\linewidth]{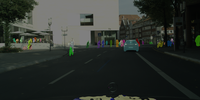} &
\includegraphics[width=0.25\linewidth]{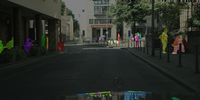} &
\includegraphics[width=0.25\linewidth]{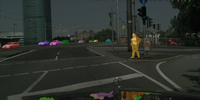}\\
\includegraphics[width=0.25\linewidth]{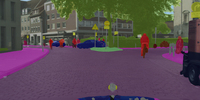} &
\includegraphics[width=0.25\linewidth]{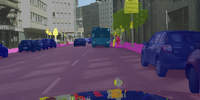} &
\includegraphics[width=0.25\linewidth]{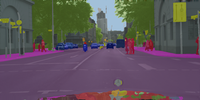} &
\includegraphics[width=0.25\linewidth]{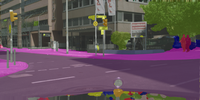}\\
\includegraphics[width=0.25\linewidth]{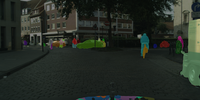} &
\includegraphics[width=0.25\linewidth]{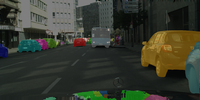} &
\includegraphics[width=0.25\linewidth]{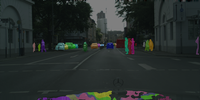} &
\includegraphics[width=0.25\linewidth]{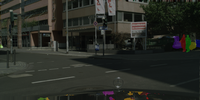}\\
\includegraphics[width=0.25\linewidth]{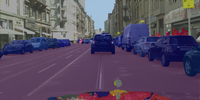} &
\includegraphics[width=0.25\linewidth]{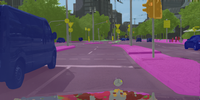} &
\includegraphics[width=0.25\linewidth]{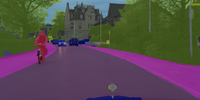} &
\includegraphics[width=0.25\linewidth]{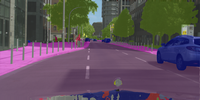}\\
\includegraphics[width=0.25\linewidth]{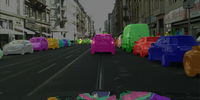} &
\includegraphics[width=0.25\linewidth]{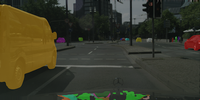} &
\includegraphics[width=0.25\linewidth]{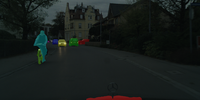} &
\includegraphics[width=0.25\linewidth]{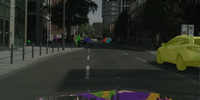}\\
\end{tabular} \egroup 
\end{center}
\caption{Visualization of Panoptic-DeepLab with Xception-71 on Cityscapes {\it val} set. Only single scale inference is used and the model achieves $63.0\%$ PQ. The first row is panoptic prediction and the second row is instance prediction.}
\label{fig:vis_cityscapes}
\end{figure*}

\begin{figure*}[!t]
\begin{center}
\bgroup 
 \def\arraystretch{0.2} 
 \setlength\tabcolsep{0.2pt}
\begin{tabular}{ccccc}
\includegraphics[width=0.2\linewidth]{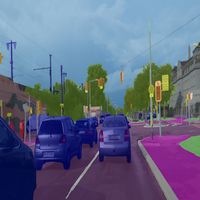} &
\includegraphics[width=0.2\linewidth]{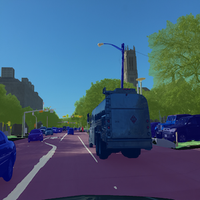} &
\includegraphics[width=0.2\linewidth]{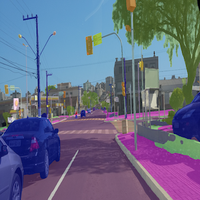} &
\includegraphics[width=0.2\linewidth]{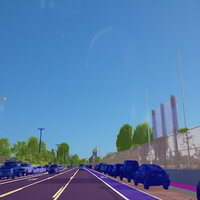} &
\includegraphics[width=0.2\linewidth]{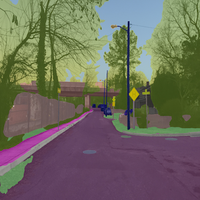}\\
\includegraphics[width=0.2\linewidth]{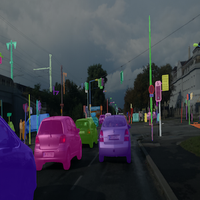} &
\includegraphics[width=0.2\linewidth]{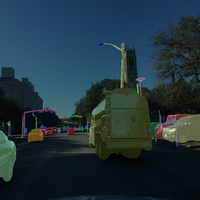} &
\includegraphics[width=0.2\linewidth]{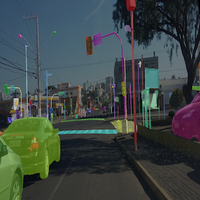} &
\includegraphics[width=0.2\linewidth]{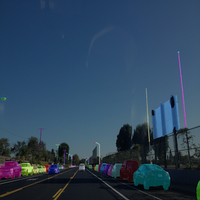} &
\includegraphics[width=0.2\linewidth]{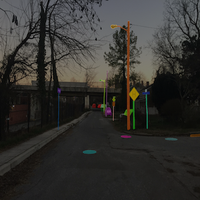}\\
\includegraphics[width=0.2\linewidth]{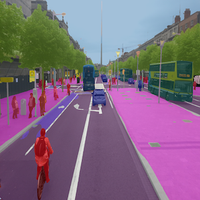} &
\includegraphics[width=0.2\linewidth]{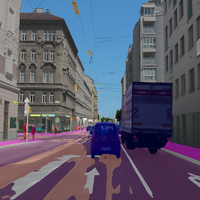} &
\includegraphics[width=0.2\linewidth]{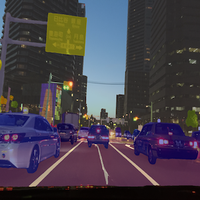} &
\includegraphics[width=0.2\linewidth]{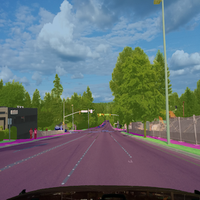} &
\includegraphics[width=0.2\linewidth]{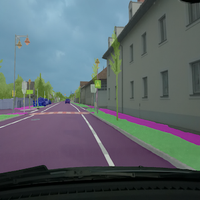}\\
\includegraphics[width=0.2\linewidth]{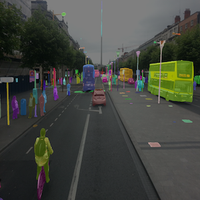} &
\includegraphics[width=0.2\linewidth]{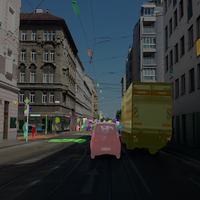} &
\includegraphics[width=0.2\linewidth]{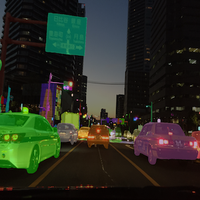} &
\includegraphics[width=0.2\linewidth]{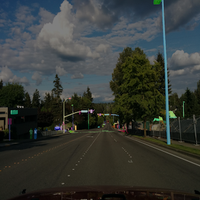} &
\includegraphics[width=0.2\linewidth]{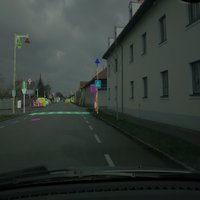}\\
\includegraphics[width=0.2\linewidth]{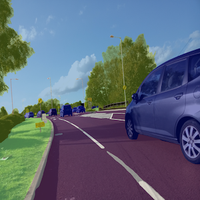} &
\includegraphics[width=0.2\linewidth]{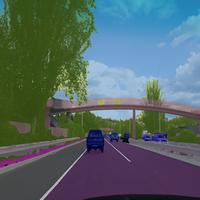} &
\includegraphics[width=0.2\linewidth]{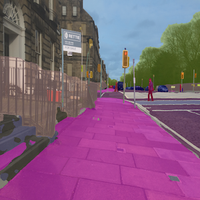} &
\includegraphics[width=0.2\linewidth]{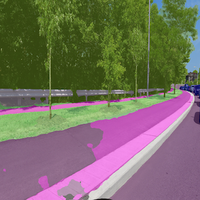} &
\includegraphics[width=0.2\linewidth]{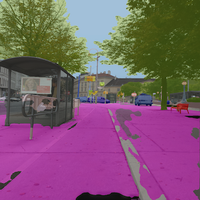}\\
\includegraphics[width=0.2\linewidth]{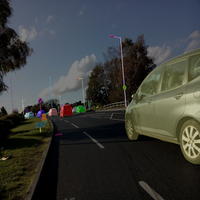} &
\includegraphics[width=0.2\linewidth]{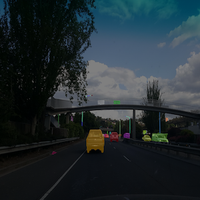} &
\includegraphics[width=0.2\linewidth]{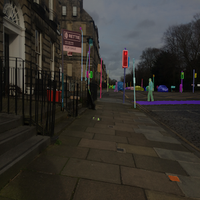} &
\includegraphics[width=0.2\linewidth]{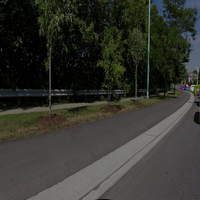} &
\includegraphics[width=0.2\linewidth]{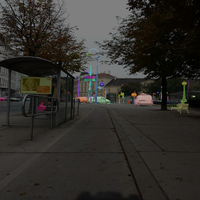}\\
\end{tabular} \egroup 
\end{center}
\caption{Visualization of Panoptic-DeepLab with Xception-71 on Mapillary Vistas {\it val} set. Only single scale inference is used and the model achieves $37.7\%$ PQ. The first row is panoptic prediction and the second row is instance prediction.}
\label{fig:vis_mapillary}
\end{figure*}

\begin{figure*}[!t]
\begin{center}
\bgroup 
 \def\arraystretch{0.2} 
 \setlength\tabcolsep{0.2pt}
\begin{tabular}{ccccc}
\includegraphics[width=0.2\linewidth]{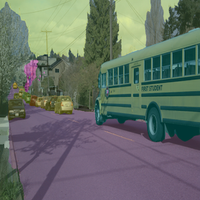} &
\includegraphics[width=0.2\linewidth]{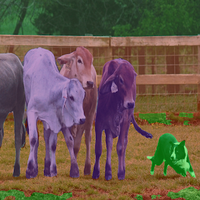} &
\includegraphics[width=0.2\linewidth]{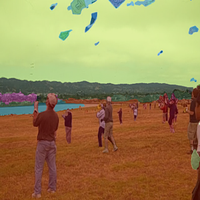} &
\includegraphics[width=0.2\linewidth]{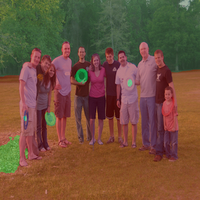} &
\includegraphics[width=0.2\linewidth]{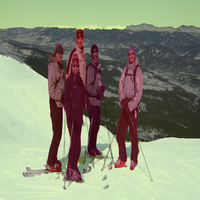}\\
\includegraphics[width=0.2\linewidth]{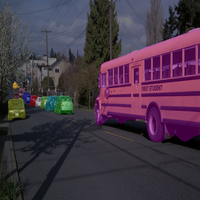} &
\includegraphics[width=0.2\linewidth]{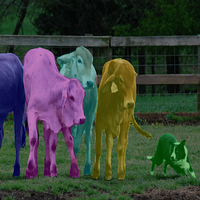} &
\includegraphics[width=0.2\linewidth]{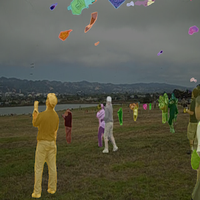} &
\includegraphics[width=0.2\linewidth]{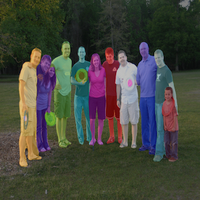} &
\includegraphics[width=0.2\linewidth]{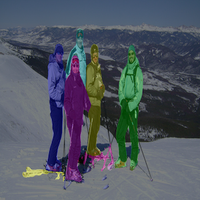}\\
\includegraphics[width=0.2\linewidth]{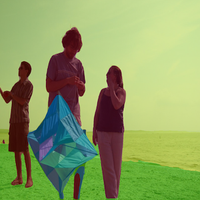} &
\includegraphics[width=0.2\linewidth]{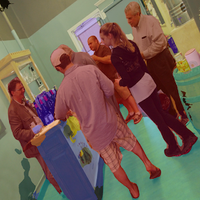} &
\includegraphics[width=0.2\linewidth]{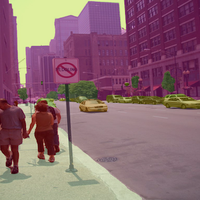} &
\includegraphics[width=0.2\linewidth]{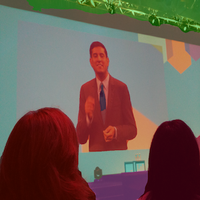} &
\includegraphics[width=0.2\linewidth]{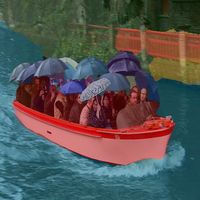}\\
\includegraphics[width=0.2\linewidth]{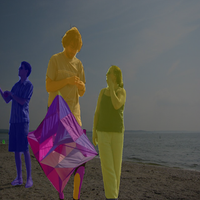} &
\includegraphics[width=0.2\linewidth]{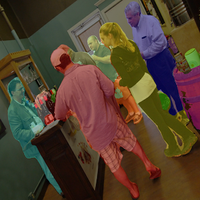} &
\includegraphics[width=0.2\linewidth]{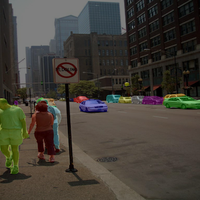} &
\includegraphics[width=0.2\linewidth]{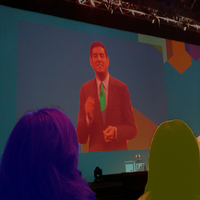} &
\includegraphics[width=0.2\linewidth]{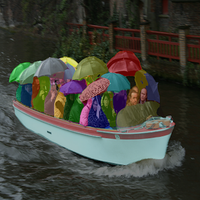}\\
\includegraphics[width=0.2\linewidth]{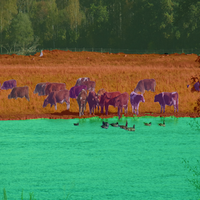} &
\includegraphics[width=0.2\linewidth]{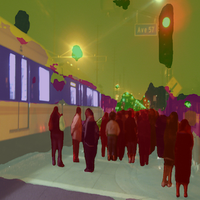} &
\includegraphics[width=0.2\linewidth]{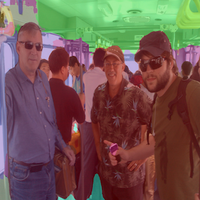} &
\includegraphics[width=0.2\linewidth]{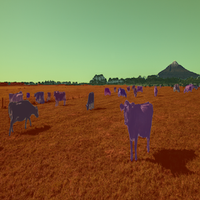} &
\includegraphics[width=0.2\linewidth]{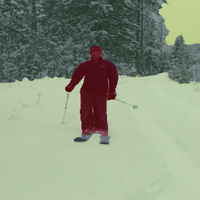}\\
\includegraphics[width=0.2\linewidth]{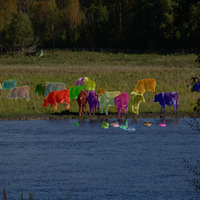} &
\includegraphics[width=0.2\linewidth]{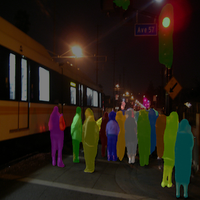} &
\includegraphics[width=0.2\linewidth]{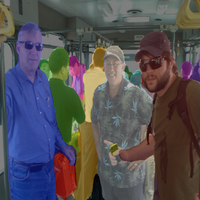} &
\includegraphics[width=0.2\linewidth]{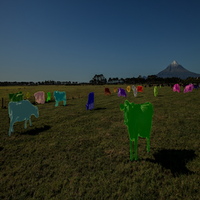} &
\includegraphics[width=0.2\linewidth]{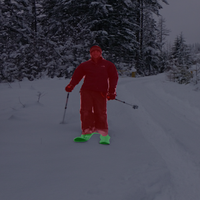}\\
\end{tabular} \egroup 
\end{center}
\caption{Visualization of Panoptic-DeepLab with Xception-71 on COCO {\it val} set. Only single scale inference is used and the model achieves $39.7\%$ PQ. The first row is panoptic prediction and the second row is instance prediction.}
\label{fig:vis_coco}
\end{figure*}

We provide more visualization results of our Panoptic-DeepLab in \figref{fig:vis_cityscapes}, \figref{fig:vis_mapillary}, and \figref{fig:vis_coco}.

{\small
\bibliographystyle{ieee_fullname}
\bibliography{egbib}
}

\end{document}